\documentclass{techreport}

\renewcommand{\reportshorttitle}{vSFT: Learning to Verbally Tune Agents at Test Time}
\renewcommand{\reportdate}{\today}
\newcommand{\reportorganization}{DreamX Team}
\newcommand{\reportpdftitle}{vSFT: Learning to Verbally Tune Agents at Test Time}
\newcommand{\reportpdfauthor}{Qishuai Wen, Rui Chen, Haiwen Li, Zhiyuan Huang, Qiwen Gu, Bingjie Gao, Geng Li, Jing Tang, Xiangxiang Chu, Chun-Guang Li}
\newcommand{\reportpdfsubject}{Verbally Supervised Fine-Tuning for LLM agents}
\newcommand{\reportpdfkeywords}{vSFT, LLM agents, test-time adaptation, hypernetworks, LoRA}

\setreportlogo{assets/dreamx_logo_report_blue_x_wordmark.png}

\title{vSFT: Learning to Verbally Tune Agents at Test Time}
\author[1, \dagger]{Qishuai Wen}
\author[1, \ddagger]{Rui Chen}
\author[1]{Haiwen Li}
\author[2]{Zhiyuan Huang}
\author[1,3]{Qiwen Gu}
\author[1,4]{Bingjie Gao}
\author[1]{Geng Li}
\author[1, \S]{Jing Tang}
\author[1]{Xiangxiang Chu}
\author[5, \S]{Chun-Guang Li}
\affiliation[1]{AMAP, Alibaba Group}
\affiliation[2]{Shenzhen Loop Area Institute}
\affiliation[3]{Tongji University}
\affiliation[4]{Shanghai Jiao Tong University}
\affiliation[5]{Independent Researcher}
\contribution[\ddagger]{Project lead}
\contribution[\S]{Corresponding authors}
\abstract{Large language model agents are increasingly improved via agentic reinforcement learning, which can be on-policy but suffers from expensive rollouts and sparse rewards, limiting its application at test time.
In this paper, we present a \textbf{v}erbally \textbf{S}upervised \textbf{F}ine-\textbf{T}uning (vSFT) framework that enables the conjunction of dense feedback in natural language and weight updates at test time.
Specifically, our vSFT forms a closed loop between an agent and a supervisor via a hypernetwork, in which the supervisor generates verbal supervision from the agent's failed trajectories, and the hypernetwork converts that supervision into low-rank weight updates to tune the agent.
We validate that this test-time in-parameter evolving process outperforms both its in-context counterparts and prevalent reinforcement learning methods.
In particular, our vSFT using Qwen3-4B outperforms GRPO and Reflexion by 8.7\% and 13.9\%
in success rate on ALFWorld, and by 4.8\% and 18.2\% on WebShop, respectively.
We believe that learning how to adapt a model, rather than an adapted model, offers a promising path towards more capable agents that improve themselves.

}
\metadata[GitHub]{\url{https://github.com/AMAP-ML/vSFT}}
\correspondence{Jing Tang: \email{guangyu.tj@alibaba-inc.com};
                Chun-Guang Li: \email{lichunguang@bupt.edu.cn}}

\usepackage{amsmath}
\usepackage{amssymb}
\usepackage{array}
\usepackage{colortbl}
\usepackage{enumitem}
\usepackage{fancyhdr}
\usepackage{float}
\usepackage{xurl}
\usepackage{algorithm}
\usepackage{algorithmic}
\usepackage{pifont}
\usepackage{multirow}
\usepackage{arydshln}
\usepackage{wrapfig}
\usepackage{needspace}
\usepackage{capt-of}
\usepackage{tabularx}
\usepackage[normalem]{ulem}
\usepackage{textcomp}
\usepackage{listings}
\tcbuselibrary{skins,breakable,listings}

\newcommand{\cmark}{\textcolor{green!70!black}{\ding{51}}}
\newcommand{\xmark}{\textcolor{red!80!black}{\ding{55}}}
\newcommand{\myparagraph}[1]{\noindent\textbf{#1.}}
\newcommand{\best}[1]{\cellcolor{bestcolor}#1}
\newcommand{\second}[1]{\cellcolor{secondcolor}#1}
\newcommand{\third}[1]{\cellcolor{thirdcolor}#1}
\definecolor{bestcolor}{RGB}{248, 215, 218}
\definecolor{secondcolor}{RGB}{250, 224, 170}
\definecolor{thirdcolor}{RGB}{222, 235, 247}
\DeclareRobustCommand{\bestlegend}[1]{\begingroup\setlength{\fboxsep}{1pt}\colorbox{bestcolor}{#1}\endgroup}
\DeclareRobustCommand{\secondlegend}[1]{\begingroup\setlength{\fboxsep}{1pt}\colorbox{secondcolor}{#1}\endgroup}
\DeclareRobustCommand{\thirdlegend}[1]{\begingroup\setlength{\fboxsep}{1pt}\colorbox{thirdcolor}{#1}\endgroup}

\definecolor{VSFTPromptBorder}{HTML}{68717B}
\newcommand{\vsftpromptfont}{\fontencoding{T1}\fontfamily{lmtt}}
\ifdefined\XeTeXversion
  \renewcommand{\vsftpromptfont}{\fontencoding{TU}\fontfamily{lmtt}}
\fi
\lstdefinestyle{vsftsystemliteral}{
  language={},
  basicstyle=\vsftpromptfont\fontseries{m}\fontshape{n}\fontsize{9}{10.5}\selectfont\color{black},
  keywordstyle={},
  commentstyle={},
  stringstyle={},
  identifierstyle={},
  columns=fullflexible,
  keepspaces=true,
  showspaces=false,
  showtabs=false,
  showstringspaces=false,
  upquote=true,
  breaklines=true,
  breakatwhitespace=true,
  breakautoindent=true,
  breakindent=2em,
  tabsize=2,
  numbers=none,
  mathescape=false,
  texcl=false,
  aboveskip=0pt,
  belowskip=0pt,
  literate={-}{{\char45\relax}}1
           {–}{{\textendash}}1
           {—}{{\textemdash}}1
}
\newtcblisting[auto counter,number within=section]{vsftsystemprompt}[2][]{
  enhanced,
  breakable,
  listing only,
  listing options={style=vsftsystemliteral},
  width=\linewidth,
  colback=white,
  colframe=VSFTPromptBorder!50!white,
  colbacktitle=VSFTPromptBorder!9!white,
  coltitle=black,
  fonttitle=\small\sffamily\bfseries,
  boxrule=0.35pt,
  arc=0.7mm,
  boxsep=0pt,
  left=7pt,
  right=7pt,
  top=7pt,
  bottom=7pt,
  toptitle=5pt,
  bottomtitle=5pt,
  before skip=8pt,
  after skip=8pt,
  pad at break*=1.5mm,
  title={Prompt~\thetcbcounter: #2},
  title after break={Prompt~\thetcbcounter: #2 (continued)},
  #1
}

\definecolor{BrandPurple}{HTML}{7366CC}
\definecolor{BrandCyan}{HTML}{00B4E5}
\definecolor{LightGray}{HTML}{F5F5F5}
\definecolor{TableRowAlt}{HTML}{EAF4FB}
\definecolor{HeaderGray}{HTML}{888888}
\definecolor{TableRule}{HTML}{AEBBC5}

\newcolumntype{L}[1]{>{\raggedright\arraybackslash}m{#1}}
\newcolumntype{C}[1]{>{\centering\arraybackslash}m{#1}}

\newtcolorbox{highlight}{
    colback=BrandCyan!5,
    colframe=BrandCyan,
    arc=2pt,
    boxrule=0.8pt,
    left=6pt, right=6pt, top=4pt, bottom=4pt,
    breakable
}

\newtcolorbox{keyfinding}{
    colback=BrandPurple!5,
    colframe=BrandPurple,
    coltitle=white,
    fonttitle=\bfseries,
    title=Key Finding,
    arc=2pt,
    boxrule=1pt,
    left=6pt, right=6pt, top=4pt, bottom=4pt,
    breakable
}

\newtcolorbox{tipbox}[1]{
    colback=ReportSurface,
    colframe=ReportPrimary,
    coltitle=white,
    fonttitle=\bfseries,
    title=#1,
    arc=2pt,
    boxrule=1pt,
    left=6pt, right=6pt, top=4pt, bottom=4pt,
    breakable
}

\fancypagestyle{plain}{%
    \fancyhf{}
    \fancyhead[L]{\footnotesize\color{HeaderGray}\sffamily \reportshorttitle}
    \fancyhead[R]{\footnotesize\color{HeaderGray} \thepage}
    \fancyfoot[C]{}

}

\fancypagestyle{firstpage}{%
    \fancyhf{}
    \fancyhead[L]{\small\sffamily\color{ReportPrimary} \reportorganization}
    \fancyhead[R]{\small\color{HeaderGray} \reportdate}
    \fancyfoot[C]{\footnotesize\color{gray}\thepage}

}

\def\ie{i.e.}
\def\eg{e.g.}
\def\th{\boldsymbol{\theta}}
\def\ph{\boldsymbol{\phi}}

\def\a{\boldsymbol{a}}
\def\c{\boldsymbol{c}}

\def\h{\boldsymbol{h}}
\def\o{\boldsymbol{o}}
\def\p{\boldsymbol{p}}
\def\q{\boldsymbol{q}}
\def\x{\boldsymbol{x}}

\def\z{\boldsymbol{z}}

\def\A{\boldsymbol{A}}
\def\B{\boldsymbol{B}}
\def\W{\boldsymbol{W}}

\def\agent{{\boldsymbol{\pi}}_{\th}}
\def\param{\th}
\def\super{\mathcal{S}}
\def\hyper{\h_{\ph}}

\def\sys{\p}
\def\prompt{\x}
\def\promptref{\tilde{\x}}
\def\condi{\c}

\def\few{\mathcal{F}}

\def\trajref{\tilde{\boldsymbol\tau}}

\def\fifo{B}
\def\numtrial{T}

\newcommand{\agentt}[1]{\boldsymbol\pi_{\th^{(#1)}}}
\newcommand{\paramt}[1]{\th^{(#1)}}
\newcommand{\memt}[1]{\mathcal{M}^{(#1)}}
\newcommand{\condit}[1]{\c^{(#1)}}
\newcommand{\trajt}[1]{\boldsymbol\tau^{(#1)}}
\newcommand{\outcomet}[1]{y^{(#1)}}
\newcommand{\feedbackt}[1]{\z^{(#1)}}

\hypersetup{
  pdftitle={\reportpdftitle},
  pdfauthor={\reportpdfauthor},
  pdfsubject={\reportpdfsubject},
  pdfkeywords={\reportpdfkeywords}
}
\begin{document}
\maketitle
\begingroup
\renewcommand{\thefootnote}{$\dagger$}
\footnotetext{Work done during internship at AMAP, Alibaba Group}
\endgroup
\thispagestyle{firstpage}
\begin{figure}[H]
\centering
\includegraphics[width=0.95\textwidth]{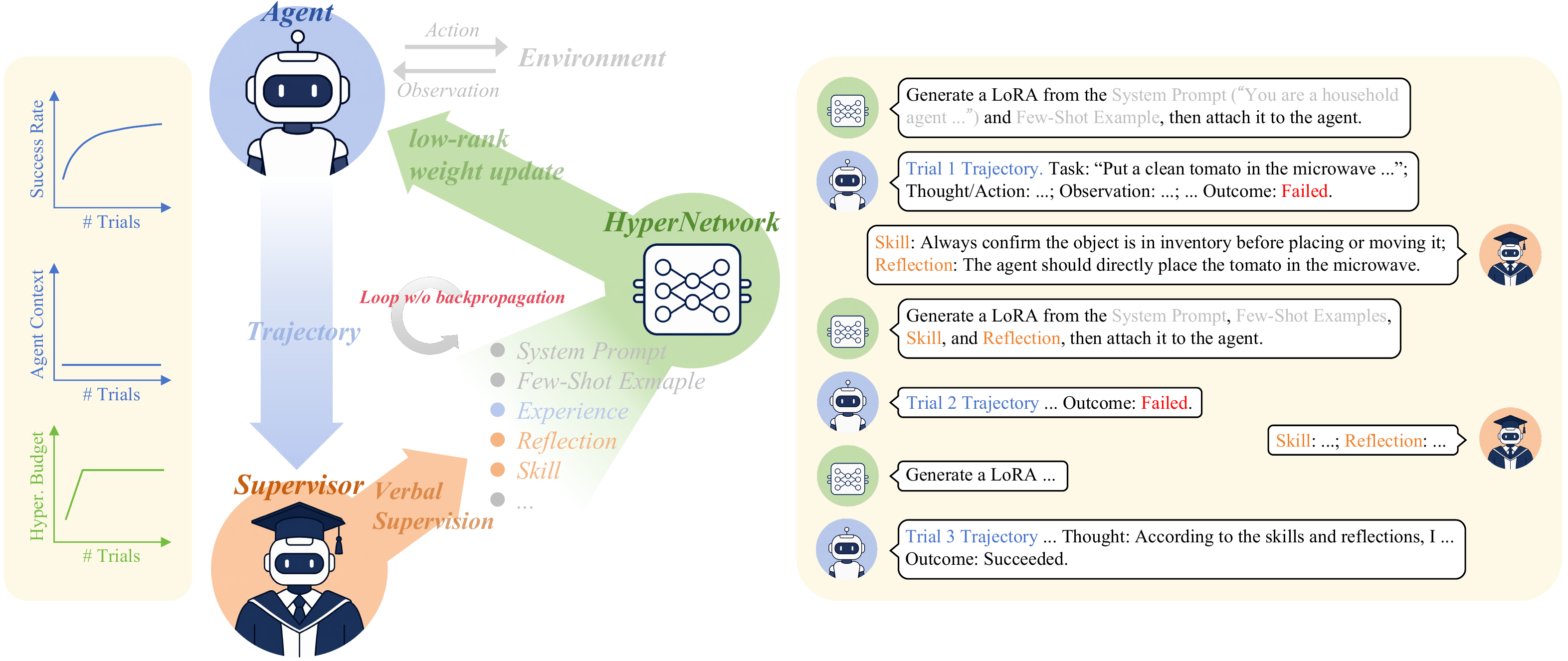}
\vspace{-0.6em}
\caption{
\small
\myparagraph{Overview of our proposed vSFT}
Given a task, 
the hypernetwork first internalizes the system prompt and few-shot examples into an initial LoRA adapter, 
and then progressively incorporates the supervisor's feedback through subsequent trial-and-error iterations,
thereby adapting the agent over time.
}
\label{fig:vSFT}
\end{figure}

\section{Introduction}
\label{sec:intro}
Modern autonomous agents use large language models (LLMs) as policies for interactive decision-making, enabling them to plan, act, call tools, and perceive the environment over multiple turns~\citep{yao2023react}.
Although pretrained LLMs possess strong general-purpose capabilities, they are not natively optimized for specific agent tasks, requiring to be further adapted via post-training methods for parameters updating~\citep{jiang2025adaptation}.

Recently, supervised fine-tuning (SFT) and reinforcement learning (RL) have been widely used for post-training LLM agents~\citep{zhang2025landscape,ji2026tree}.
In particular, on-policy distillation (OPD) becomes a promising paradigm~\citep{wang2026skill,lu2026self}.
Given golden trajectories, SFT is simple and data-efficient for forcing token-level imitation across positions in parallel, however, it is vulnerable to distribution shift on unseen trajectories due to the inherent off-policy nature~\citep{nie2026post}.
Moreover, it could also be problematic to collect golden trajectories for open-world agent tasks.
In contrast, RL is on-policy, evolving the agent from its own rollouts rather than imitating from a fixed trajectory dataset.
Nonetheless, RL often relies on sparse rewards at trajectory level and GRPO-style methods~\citep{shao2024deepseekmath,feng2026group,chu2026gpg} which further increase the rollout costs due to multiple sampled trajectories.

\begin{wrapfigure}[14]{r}{0.35\textwidth}
\vspace{-11pt}
\centering
\hspace{-1.8mm}
\includegraphics[width=0.26\columnwidth]{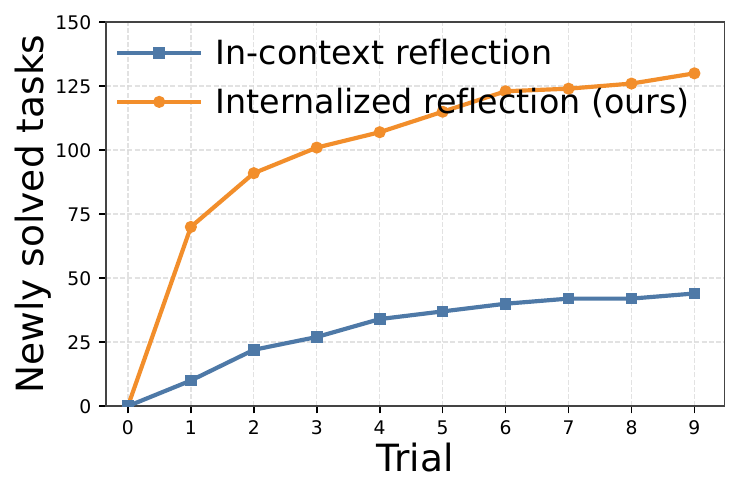}
\vspace{-1em}
\caption{
\small
\myparagraph{Efficacy comparison}
On the shared set of tasks that both methods fail in the initial trial,
we replace the in-context reflections with our internalized reflections without altering their content
to fairly compare the effectiveness in guiding the agent in a controlled setting.
}
\label{fig:motivation}
\end{wrapfigure}

As a lightweight alternative to RL fine-tuning, \citet{shinn2023reflexion} introduced a verbal reinforcement learning (vRL), which reinforces agents at test time through in-context reflections.
Unlike post-training methods that directly update weights and thereby improve the agents essentially, however, vRL relies on in-context learning, which incurs recurring context overhead and may become less effective in long-horizon tasks.\footnote{
Even LLMs that have undergone instruction fine-tuning can struggle to reliably satisfy constraints specified in context~\citep{zhang2026firebench}. 
For example, prompting alone does not guarantee that generated JSON is syntactically valid~\citep{Lu0CZWLLL025}.}
Similar to prior work~\citep{zhao2024context,geng2026control}, we observe that, regardless of whether the reflections could help complete the task, the agent often struggles to follow them, suggesting that in-context reflections cannot reliably steer the agent's behavior (as shown in Fig.~\ref{fig:motivation}).
This motivates us to develop an end-to-end trainable method that learns to internalize verbal feedback into model parameters to reduce context overhead and enable more reliable behavioral adaptation.

Inspired by recent works in hypernetwork-generated LoRA adapters that convert natural language into low-rank weight updates~\citep{charakorn2025text,charakorn2026doc}, in this paper, we propose \textbf{vSFT}, a \textbf{v}erbally \textbf{S}upervised \textbf{F}ine-\textbf{T}uning framework for tuning agents at test time via verbal supervision, as illustrated in Fig.~\ref{fig:vSFT}.
To be specific, in vSFT, we build an iterative closed loop among agent, environment and supervisor via a trained lightweight hypernetwork, in which the agent interacts with the environment, the supervisor model provides verbal supervision (\ie, verbal feedback, such as skills and reflections), and the hypernetwork converts the verbal supervision into weight updates in the form of a LoRA adapter to tune the agent.
Here, the hypernetwork serves as a parameterized, learnable interface through which the agent iteratively improves itself in response to self-reflection or external feedback, supplying a mechanism pretrained LLMs otherwise lack~\citep{zweiger2026self}.

\begin{table}[ht]
\centering
\small
\caption{
\small
A summary of existing methods for improving LLM agents with respect to four desirable properties.
Our vSFT is the only method that satisfies all of them.
As discussed in \S\ref{sec:vsft}, vSFT can be viewed as an extension of OPD that requires neither access to next-token distributions nor backpropagation once the hypernetwork has been trained, thereby enabling test-time adaptation under supervision from closed-source API models.
}
\vspace{-1em}
\begin{tabular}{lcccc}
\toprule
Method & On policy & In parameter & Dense feedback & W/o backprop. \\
\midrule
SFT & \xmark & \cmark & \cmark & \xmark \\
RL & \cmark & \cmark & \xmark & \xmark \\
OPD & \cmark & \cmark & \cmark & \xmark \\
vRL~\citep{shinn2023reflexion} & \cmark & \xmark & \cmark & \cmark \\
JitRL~\citep{li2026just} & \cmark & \xmark & \xmark & \cmark \\
\rowcolor{cyan!10}
vSFT & \cmark & \cmark & \cmark & \cmark \\
\bottomrule
\end{tabular}
\label{tab:methods}
\end{table}

As summarized in Tab.~\ref{tab:methods}, our vSFT is the only method that satisfies all of the four desired properties: a) on-policy (\ie, at test-time) improvement from the agent's own rollouts, reducing the train-test mismatch; b) in-parameter adaptation rather than in-context learning, enabling more reliable and context-efficient steering\footnote{Human intelligence and self-improvement are not driven by an ever-growing context but through the changes in neural connections (e.g., Hebbian learning), much as weight updates in neural networks.}; c) using dense feedback, mitigating the credit assignment problem in agentic RL~\citep{zhang2026reasoning}; and d) backpropagation-free test-time fine-tuning, rather than being restricted to offline post-training.

In experiments, we find that our vSFT outperforms both its in-context counterparts and prevalent reinforcement learning methods.
In particular, our vSFT using Qwen3-4B outperforms GRPO and Reflexion by 11.5\% and 10.8\% in success rate on ALFWorld's in-distribution split, and by 4.8\% and 18.2\% on WebShop, respectively, while using only 4\% of GRPO's training rollout budget on ALFWorld and 10\% on WebShop.
Moreover, when further using Qwen-Plus as the supervisor, our vSFT substantially outperforms Qwen-Plus and performs on par with Qwen3.8-Max, despite the agent having under 5\% as many active parameters.
We further show that these improvements on individual tasks can be consolidated into a persistent parametric adaptation via LoRA merging, which generalizes to new tasks.

\myparagraph{Paper contribution}
The main contributions of the paper are highlighted as follows.
\begin{enumerate}[leftmargin=*, topsep=0.25em, noitemsep]
\item We propose vSFT, a novel framework that iteratively improves an agent at test time through a closed loop among the agent's behavior, the supervisor's verbal feedback, and the weight updates to the agent, thereby enabling test-time parameter adaptation (\S\ref{sec:vsft}).
\item We provide an effective and parameterized mechanism for agent self-improvement by using a trained hypernetwork to iteratively internalize verbal feedback into the agent's parameters without requiring back-propagation (\S\ref{sec:hypernetwork}).
\item We empirically validate the effectiveness of our vSFT in self-improvement and exhibit its strong generalization ability and robustness (\S\ref{sec:experiments}).
\end{enumerate}

\section{Related work}

\subsection{LLM agents and their adaptation}
Large language models (LLMs) have enabled the development of autonomous agents by serving as policies for interactive decision-making.
Nonetheless, even capable pretrained LLMs often require further adaptation to specific agentic tasks.
While such adaptation can target either the agent itself or other callable components, in this paper, we focus on adapting the agent itself.

Existing adaptation methods are mainly based on post-training (i.e., fine-tuning).
Supervised fine-tuning (SFT) suffers from limited generalization due to its off-policy nature~\citep{chen2025retaining,shenfeld2026rl}, whereas reinforcement learning (RL) can be deployed on-policy but demands expensive rollouts.
On the other hand, lightweight test-time alternatives have recently attracted increasing attention.
For example, verbal reinforcement learning (vRL)~\citep{shinn2023reflexion,kim2025reflact,li2026m2note} adapts agents through verbal reflections and in-context learning rather than scalar rewards and in-parameter updates; MemRL~\citep{zhang2026memrl} performs reinforcement learning over episodic memory; and just-in-time reinforcement learning (JitRL)~\citep{li2026just} estimates action advantages from past experiences.
Unlike these approaches that operate in context, memory space, or action space,
our vSFT bridges verbal feedback and parameter adaptation through a hypernetwork.
Similar to on-policy distillation (OPD)~\citep{agarwal2024policy}, vSFT can be viewed as an on-policy extension of SFT.
However, OPD is typically applied during post-training, whereas vSFT runs its closed loop at test time and can therefore complement post-training methods.

\subsection{Hypernetwork-generated LoRAs}
Hypernetwork-generated LoRAs (HyperLoRAs) have recently emerged as a promising approach for instant task adaptation~\citep {charakorn2025text} and knowledge injection~\citep{charakorn2026doc,liu2026shine}, in which textual context (\eg, task descriptions or documents) is converted into LoRA adapters by a specially trained hypernetwork.
Such a context-to-parameter paradigm offers two key advantages: a) adapters can be represented, stored, and edited in natural language, while being instantiated as LoRAs only when needed; and b) new adapters can be generated for unseen tasks (or documents) instantly without additional training.
Soon then, follow-up work extends HyperLoRAs to repository internalization for code language models~\citep{hotsko2026code2lora} and to skill internalization for agents~\citep{yu2026latentskill}.
Our vSFT instead explores a new application for HyperLoRA, \ie, using it as a learnable component that enables an agent to iteratively update its weights at test time in response to verbal feedback from a supervisor.
\section{Methods}
\label{sec:methods}
In this section, we present the proposed vSFT framework, which consists of two separate parts: a) \emph{test-time in-parameter agent adaptation} (\S\ref{sec:vsft}), which is a closed-loop process in which the hypernetwork iteratively adapts the agent by converting verbal feedback from a supervisor, who diagnoses the agent's failures, into weight updates to the agent; and (b) \emph{offline hypernetwork meta-training} (\S\ref{sec:hypernetwork}), in which the hypernetwork is trained to generate a LoRA adapter that effectively steers the agent's behavior.
In addition, we also discuss the connections between vSFT and OPD and the potentials of the hypernetwork to serve as a back-propagation-free optimizer.

\subsection{Test-time in-parameter agent adaptation}
\label{sec:vsft}

\subsubsection{Preliminaries and notation}
We denote an agent as $\agent$, with $\th$ denoting its pretrained parameters.
The agent attempts to solve a task specified by a task instruction $\prompt$ (\eg, ``\emph{put a clean tomato in the microwave}'' in a household setting), given a system prompt $\sys$ and few-shot examples $\few$.
The agent is allowed up to $T$ trials until success.
After a failed trial, a supervisor $\super$ diagnoses the trajectory and provides verbal feedback to guide the agent in the next trial.
Unlike vRL~\citep{shinn2023reflexion}, where verbal feedback is injected into the agent's context while its parameters remain unchanged across trials, our vSFT employs a hypernetwork $\hyper$ to internalize verbal feedback into the agent's parameters, updating them across trials.
We therefore denote the agent at trial $t$ as $\agentt{t}$, where $\paramt{t}$ denotes its updated parameters.

\subsubsection{Closed loop among agent, supervisor, and hypernetwork}
Specifically, our vSFT constructs a closed loop at test time among the agent, the supervisor, and the hypernetwork, as illustrated in Fig.~\ref{fig:vSFT}.
Next, we will introduce the three components of our vSFT and describe how they interact with each other within the loop.

\myparagraph{Agent}
The agent at trial $t$ interacts with the environment over multiple steps, yielding a trajectory
\begin{equation}
  \trajt{t} = (\o^{(t)}_1,\a^{(t)}_1, \o^{(t)}_2,\a^{(t)}_2,\ldots,\o^{(t)}_{s_t},\a^{(t)}_{s_t}, y^{(t)}),
\end{equation}
where $\o^{(t)}_s$ and $\a^{(t)}_s$ denote the observation and the action at the $s$-th interaction step, respectively, and $\outcomet{t}$ denotes the final outcome (\eg, success or failure).
An action is sampled from the current policy $\agentt{t}$ as
\begin{equation}
  \a^{(t)}_s \sim \boldsymbol\pi_{\th^{(t)}} \left(\cdot \mid \x, \o^{(t)}_{\leq s}, \a^{(t)}_{<s}\right),
  \label{eq:action}
\end{equation}
where agent $\agentt{t}$ observes only the task instruction and the within-trial interaction history up to $s$.
That is, in vSFT, the system prompt, few-shot examples, skills, reflections, and other forms of memory are internalized into $\paramt{t}$, rather than being injected into the agent's context across trials.

\myparagraph{Supervisor}
The supervisor $\super$ can be either the agent itself or a more capable model.
If trial $t$ fails, the supervisor diagnoses the failed trajectory and generates the verbal feedback $\feedbackt{t}$ as
\begin{equation}
  \feedbackt{t} = \super(\sys, \prompt, \trajt{t}, \promptref, \trajref, \memt{t}),
  \label{eq:verbal_feedback}
\end{equation}
where $\promptref$ and $\trajref$ denote a successfully solved task and its corresponding trajectory, respectively, and $\memt{t}$ denotes the accumulated verbal feedback (\ie, memory) from previous trials.
The verbal feedback essentially consists of \emph{a skill} and \emph{a reflection}---skill encodes reusable procedural knowledge that is transferable across tasks and reflection diagnoses what went wrong in the current trajectory.
Following vRL~\citep{shinn2023reflexion}, we maintain the accumulated verbal feedback $\memt{t}$ in a ``\emph{first-in, first-out}'' (FIFO) queue, \ie,
\begin{equation}
  \memt{t+1} = \operatorname{FIFO}_{\fifo}(\memt{t},\feedbackt{t}),\ \ \text{with} \ \ \memt{0}=\emptyset,
\end{equation}
where $\operatorname{FIFO}_{\fifo}$ appends $\feedbackt{t}$ to $\memt{t}$ while retaining only the most recent $\fifo$ entries.

\myparagraph{Hypernetwork}
We refer to the hypernetwork input, \ie, the textual content to be internalized into the agent's parameters by the hypernetwork as the \emph{condition}.\footnote{Since our vSFT is compatible with different base models and their associated hypernetworks (\eg, D2L~\citep{charakorn2026doc} and SHINE~\citep{liu2026shine}), we abstract away the specific architectural and implementation details of the hypernetwork and characterize it solely by its input and output.}
Specifically, we define the condition for trial $t$ as
\begin{equation}
  \condit{t} = (\sys, \few, \memt{t}).
  \label{eq:condition}
\end{equation}
Then, given the condition $\condit{t}$, the hypernetwork $\hyper$ generates a trial-specific LoRA adapter, \ie,
\begin{equation}
  \left\{ \left(\A_{k}^{(t)}, \B_{k}^{(t)}\right) \right\}_{k\in\Omega} = \hyper(\condit{t}),
  \label{eq:hypernet_adapter}
\end{equation}
where $\Omega$ denotes the index set of target weight matrices.
For each $k\in\Omega$, let $\W_k\in\mathbb{R}^{d_k^{\mathrm{out}}\times d_k^{\mathrm{in}}}$ denote the corresponding weight matrix, $\A_k^{(t)}\in\mathbb{R}^{r\times d_k^{\mathrm{in}}}$ and $\B_k^{(t)}\in\mathbb{R}^{d_k^{\mathrm{out}}\times r}$ denote the generated low-rank factor matrices.
We then induce the effective weight update for trial $t$ as follows:
\begin{equation}
  \W_{k}^{(t)} = \W_{k} + \alpha \B_{k}^{(t)}\A_{k}^{(t)},
  \label{eq:lora_effective_weight}
\end{equation}
where $\alpha$ is a scaling coefficient.
While the agent's pretrained parameters $\param$ remain unchanged throughout test time, the generated adapter changes across trials.
In our implementation, each newly generated adapter replaces the previous one rather than being appended to it.
Thus, the effective agent parameters at trial $t$ can be written as
\begin{equation}
  \paramt{t} = \param \oplus \hyper(\condit{t}), \ \ \forall t \in \{0,1,\ldots,T-1\},
  \label{eq:update}
\end{equation}
where $\oplus$ denotes equipping the generated adapter.

\begin{algorithm}[t]
\caption{\textbf{verbally Supervised Fine-Tuning (vSFT)}}
\label{alg:vsft}
\textbf{Models}: base agent $\agent$, supervisor $\super$, hypernetwork $\hyper$ \\
\textbf{Input}: system prompt $\sys$, few-shot examples $\few$, current task instruction $\prompt$, reference task instruction $\promptref$ and its trajectory $\trajref$ \\
\textbf{Hyperparameters}: FIFO size $\fifo$, maximum number of trials $\numtrial$ \\
\vspace{-1em}
\begin{algorithmic}[1]
\STATE Initialize the accumulated verbal feedback (\ie, memory): $\memt{0}\leftarrow\emptyset$.
\FOR{$t=0$ to $\numtrial-1$}
\STATE Compose the condition (\ie, hypernetwork input) for this trial: $\condit{t}\leftarrow(\sys,\few,\memt{t})$.
\STATE Internalize the condition into the agent's parameters: $\paramt{t}\leftarrow\param\oplus\hyper(\condit{t})$.
\STATE Roll out $\agentt{t}$ on $\prompt$ to obtain trajectory $\trajt{t}$ and outcome $\outcomet{t}\in\{\textsc{success},\textsc{failure}\}$.
\IF{$\outcomet{t}=\textsc{success}$}
\RETURN $\trajt{t}$
\ELSE
\STATE Diagnose the failure and generate verbal feedback: $\feedbackt{t}\leftarrow \super(\sys,\prompt,\trajt{t},\promptref,\trajref,\memt{t})$.
\STATE Update memory with a FIFO queue of size $\fifo$: $\memt{t+1}\leftarrow \operatorname{FIFO}_{\fifo}(\memt{t},\feedbackt{t})$.
\ENDIF
\ENDFOR
\RETURN $\trajt{\numtrial-1}$
\end{algorithmic}
\end{algorithm}

\myparagraph{Discussion}
Intriguingly, we find that vSFT has a close conceptual connection to OPD.
To be specific, both of vSFT and OPD mitigate the off-policy limitation of SFT by adapting the agent on states induced by its own rollouts.
Nevertheless, the distinction lies in when their adaptation occurs: OPD is during post-training, whereas vSFT is at test-time.
These two stages are complementary and can be applied in conjunction.
Moreover, both methods are distilling a teacher's (\ie, supervisor in our context) knowledge or capability into the student's parameters.
However, OPD favors consistent thinking patterns between the student and the teacher (\eg, high overlap in their top-$k$ next-token sets) for effective optimization, making it particularly sensitive to teacher--student 
distribution
mismatch~\citep{li2026rethinking,zhu2026many}.
In contrast, vSFT directly internalizes the teacher's verbal feedback, thereby avoiding explicit token-level alignment.
In a degree, our vSFT thus suggests a promising test-time extension of OPD, with a more general and interpretable supervision interface.

\begin{figure}[tb!]
\centering
\includegraphics[width=0.99\textwidth]{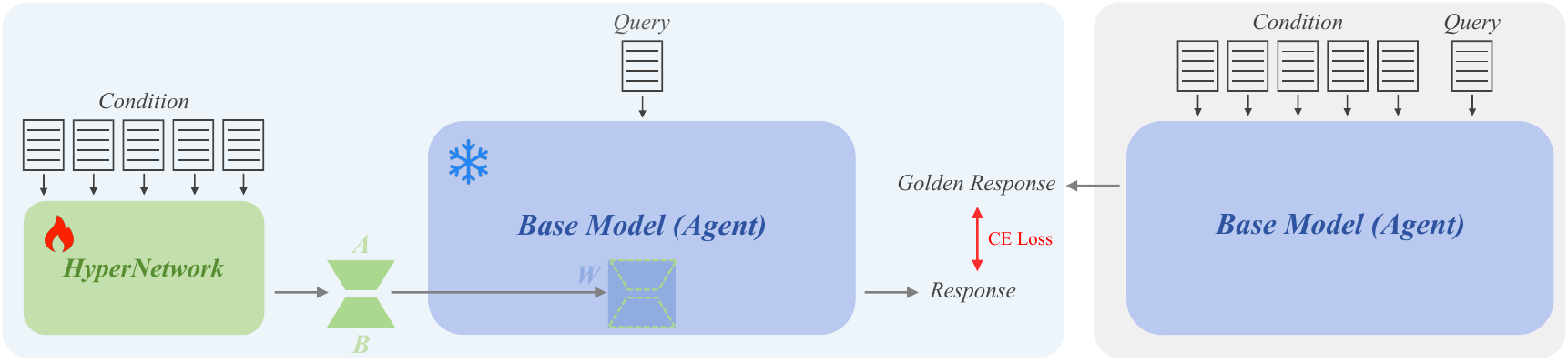}
\vspace{-1em}
\vspace{+4pt}
\caption{
\myparagraph{Hypernetwork meta-training pipeline}
The curated dataset (right panel) is essentially an SFT dataset, but consists of condition-query-response triplets rather than query-response pairs.
This design trains the hypernetwork to convert verbal feedback into LoRA adapters that effectively steer the agent's behavior (left panel).
}
\label{fig:train}
\end{figure}

\subsection{Offline hypernetwork meta-training}
\label{sec:hypernetwork}
While vSFT runs a closed loop at test time without back-propagation, the hypernetwork that bridges natural language and weight updates requires SFT-style offline training.
In another words, the conventional SFT is to \emph{directly adapt a model}, whereas the hypernetwork in our vSFT is to \emph{learn how to adapt the model}.
We therefore view this process as \emph{meta-training}, in which the self-improvement\footnote{We define self-improvement as the ability of a system (\eg, the agent and hypernetwork together in our framework) to improve its own performance (\ie, ``doing'') in response to self-reflection or external feedback, rather than merely generate high-quality reflections (\ie, ``knowing''). } mechanism is learned offline and applied at test time.
Specifically, during meta-training, the hypernetwork learns to generate LoRA adapters that enable the agent to reproduce behavior previously elicited by in-context conditions (as defined in Eq.~\ref{eq:condition}).
These conditions are now provided only to the hypernetwork rather than to the agent, as illustrated in Fig.~\ref{fig:train}.

\myparagraph{Data curation}
We first run the agent $\agent$ (alternatively another agent) in a vRL~\citep{shinn2023reflexion} framework to interact with the environment and collect its trajectories, where the conditions are directly provided in context (see the right panel of Fig.~\ref{fig:train}).
Then, we construct a dataset of triplets, each consisting of a condition, a query (prefix), and a response, i.e., $(\condi, (\prompt, \o_{\leq s}, \a_{<s}), \a_s)$, extracted from the successful trajectories:
\begin{equation}
  \mathcal{D} = \left\{ \left( \c^{i},\; \q^{i}_s,\; \a^{i}_s \right) \;\middle|\; \tau^{i} \in \mathcal{T}_{\mathrm{succ}},\; s=1,\ldots,s_i \right\},
  \label{eq:meta_dataset}
\end{equation}
where $i$ indexes completed tasks rather than trials, and $\q^i_s=(\x^i,\o^i_{\leq s},\a^i_{<s})$.
Since we retain only the final trial for each completed task, the trial index is uniquely determined by $i$ and is therefore omitted from Eq.~\ref{eq:meta_dataset}.

\myparagraph{Loss function}
During meta-training, the condition is removed from the agent's context and is provided only to the hypernetwork.
The hypernetwork then generates a LoRA adapter (as in Eqs.~\ref{eq:hypernet_adapter}, \ref{eq:lora_effective_weight}, and \ref{eq:update}), while the adapted agent aims to imitate the actions in the successful trajectory using only the task instruction and interaction history as context, with the condition internalized in its weights:
\begin{equation}
  \mathcal{L}(\boldsymbol\phi) = - \frac{1}{\sum_{i=1}^{\lvert\mathcal{T}_{\mathrm{succ}}\rvert} s_i} \sum_{i=1}^{\lvert\mathcal{T}_{\mathrm{succ}}\rvert} \sum_{s=1}^{s_i} \log \pi_{\param \oplus \hyper(\c^i)} \left( \a^i_s \mid \q_s^i \right),
  \label{eq:vsft_training_loss}
\end{equation}
where only hypernetwork parameters $\boldsymbol\phi$ are trainable.

For clarity, we summarize the procedure of our vSFT in Alg.~\ref{alg:vsft}.
Note that in vSFT the condition plays a role similar to privileged information in on-policy self-distillation (OPSD)~\citep{penaloza2026privileged}, in which OPSD learns a student policy that internalizes such privileged information through distillation, whereas our vSFT adopts a meta-learning viewpoint that learns how to transform privileged verbal feedback into low-rank weight updates.
This allows vSFT to learn adaptation from offline trajectories while performing it at test time.
From another perspective, the verbal feedback in vSFT can be viewed as a semantic counterpart of the error signal used in back-propagation.
Unlike conventional error signals derived from predefined loss functions, it is produced by an LLM and expressed in natural language.
The hypernetwork then learns to convert such verbal signals into weight updates, acting as a back-propagation-free optimizer.

\begin{table}[ht!]
\centering
\scriptsize
\setlength{\tabcolsep}{2.6pt}
\caption{
\small
\myparagraph{Comparisons on ALFWorld}
We compare vSFT against ICL and RL methods across different base models on the seen and unseen validation splits of ALFWorld, reporting success rates for six categories of tasks and the overall and averaged results.
The top three distinct results across all base models are highlighted as \bestlegend{first}, \secondlegend{second}, and \thirdlegend{third}, respectively.
A $\dagger$ indicates that Qwen-Plus is used as the supervisor for both data curation and evaluation, while a $\ddagger$ additionally indicates that the agent used for data curation (or for generating SkillRL's skills) is also Qwen-Plus
See Appendix~\ref{app:details} for a clearer illustration of their differences.
}
\vspace{-1em}
\begin{tabular}{l*{7}{c}*{7}{c}l}
\toprule
\multirow{2}{*}{Method} & \multicolumn{7}{c}{Seen} & \multicolumn{7}{c}{Unseen} & \multirow{2}{*}{Avg.} \\
\cmidrule(lr){2-8}
\cmidrule(lr){9-15}
& Pick & Look & Clean & Heat & Cool & Pick2 & Overall & Pick & Look & Clean & Heat & Cool & Pick2 & Overall & \\
\midrule
\multicolumn{16}{l}{\textcolor{black!60}{\itshape Qwen3.8-Max}} \\
\addlinespace[0.15em]
\hspace{0.5em}ReAct & \third{94.3} & \second{92.3} & \third{92.6} & \second{93.8} & 88.0 & \third{91.7} & 92.1 & 83.3 & 44.4 & 80.6 & \best{100.0} & \second{95.2} & \third{82.4} & 82.1 & 87.1 \\
\hspace{0.5em}Reflexion & \best{100.0} & \best{100.0} & \best{100.0} & \best{100.0} & \best{100.0} & \best{100.0} & \best{100.0} & \best{100.0} & \best{100.0} & \second{96.8} & \best{100.0} & \best{100.0} & \best{100.0} & \second{99.3} & \best{99.7} \\
\addlinespace[0.3em]
\multicolumn{16}{l}{\textcolor{black!60}{\itshape Qwen-Plus-2507}} \\
\addlinespace[0.15em]
\hspace{0.5em}ReAct & 77.1 & 76.9 & 51.9 & 81.2 & 80.0 & 70.8 & 72.1 & \third{91.7} & 33.3 & 64.5 & 69.6 & 81.0 & \third{82.4} & 70.9 & 71.5 \\
\hspace{0.5em}Reflexion & \best{100.0} & \second{92.3} & \second{96.3} & \best{100.0} & \second{96.0} & \best{100.0} & \third{97.9} & \best{100.0} & 72.2 & 90.3 & \third{91.3} & \best{100.0} & \best{100.0} & 92.5 & 95.2 \\
\addlinespace[0.3em]
\multicolumn{16}{l}{\textcolor{black!60}{\itshape Qwen3-32B}} \\
\addlinespace[0.15em]
\hspace{0.5em}ReAct & 71.4 & \third{84.6} & 88.9 & \second{93.8} & 84.0 & 83.3 & 82.9 & \third{91.7} & 61.1 & \third{93.5} & \second{95.7} & \third{90.5} & 64.7 & 85.1 & 84.0 \\
\hspace{0.5em}Reflexion & \second{97.1} & \best{100.0} & \best{100.0} & \best{100.0} & \third{92.0} & \second{95.8} & 97.1 & \best{100.0} & \third{88.9} & \best{100.0} & \best{100.0} & \third{90.5} & \third{82.4} & 94.8 & \third{96.0} \\
\addlinespace[0.3em]
\multicolumn{16}{l}{\textcolor{black!60}{\itshape Qwen3-14B}} \\
\addlinespace[0.15em]
\hspace{0.5em}ReAct & 80.0 & \second{92.3} & 81.5 & \third{87.5} & 64.0 & 70.8 & 77.9 & \third{91.7} & 22.2 & 77.4 & \second{95.7} & 85.7 & 52.9 & 73.9 & 75.9 \\
\hspace{0.5em}Reflexion & \second{97.1} & \second{92.3} & \best{100.0} & 93.8 & 84.0 & 83.3 & 92.1 & \best{100.0} & \second{94.4} & \third{93.5} & \best{100.0} & \second{95.2} & \third{82.4} & 94.8 & 93.5 \\
\addlinespace[0.3em]
\multicolumn{16}{l}{\textcolor{black!60}{\itshape Qwen3-4B-Instruct-2507}} \\
\addlinespace[0.15em]
\hspace{0.5em}ReAct & 68.6 & 76.9 & 63.0 & 62.5 & 64.0 & 50.0 & 63.6 & 62.5 & 66.7 & 48.4 & 43.5 & 38.1 & 47.1 & 50.7 & 57.2 \\
\hspace{0.5em}Reflexion & 82.9 & \best{100.0} & 85.2 & 75.0 & 76.0 & 79.2 & 82.1 & 75.0 & \third{88.9} & 77.4 & 73.9 & 57.1 & \third{82.4} & 75.4 & \underline{78.8} \\
\hspace{0.5em}Reflexion\smash{$^\dagger$} & 97.1 & 76.9 & \second{96.3} & 87.5 & 84.0 & 83.3 & 89.3 & 91.7 & 55.6 & 80.6 & 78.3 & \third{90.5} & 58.8 & 77.6 & \uuline{83.5} \\
\hspace{0.5em}SFT & 62.9 & 38.5 & 55.6 & 68.8 & 60.0 & 70.8 & 60.7 & 70.8 & 27.8 & 61.3 & 73.9 & 42.9 & 11.8 & 51.5 & 56.1 \\
\hspace{0.5em}SFT\,+\,Reflexion & \best{100.0} & 61.5 & \second{96.3} & \second{93.8} & \third{92.0} & 79.2 & 90.0 & \second{95.8} & 38.9 & 80.7 & 73.9 & \third{90.5} & 64.7 & 76.1 & 83.1 \\
\hspace{0.5em}GRPO & \second{97.1} & 61.5 & 85.2 & 75.0 & 80.0 & 70.8 & 81.4 & 83.3 & 77.8 & 87.1 & 87.0 & \second{95.2} & \second{88.2} & 86.6 & 84.0 \\
\hspace{0.5em}GiGPO & 91.4 & 69.2 & 66.7 & 81.3 & 84.0 & 79.2 & 80.0 & 79.2 & 66.7 & 54.8 & 78.3 & 85.7 & 70.6 & 71.6 & 75.8 \\
\hspace{0.5em}SkillRL\smash{$^\dagger$} & 82.9 & 69.2 & \second{96.3} & 75.0 & 88.0 & 87.5 & 85.0 & 58.3 & \second{94.4} & 80.6 & 73.9 & 81.0 & \second{88.2} & 73.9 & 79.5 \\
\addlinespace[0.1em]
\cdashline{1-16}
\addlinespace[0.2em]
\hspace{0.5em}\textbf{vSFT} (one trial) & 60.0 & \third{84.6} & 77.8 & \second{93.8} & 56.0 & 16.7 & 61.4 & 75.0 & 77.8 & 77.4 & \third{91.3} & 52.4 & 17.7 & 67.9 & 64.7 \\
\hspace{0.5em}\textbf{vSFT} & 97.1 & 92.3 & 96.3 & 93.8 & 84.0 & 91.7 & 92.9 & 95.8 & 88.9 & 90.3 & 100.0 & 100.0 & 76.5 & 92.5 & \underline{92.7} {\tiny(+13.9)} \\
\hspace{0.5em}\textbf{vSFT}\smash{$^\dagger$} & \best{100.0} & \best{100.0} & \best{100.0} & \second{93.8} & 88.0 & \second{95.8} & 96.4 & \best{100.0} & \second{94.4} & \best{100.0} & \best{100.0} & \best{100.0} & \third{82.4} & \third{97.0} & \second{\uuline{96.7} {\tiny(+13.2)}} \\
\hspace{0.5em}\textbf{vSFT}\smash{$^\ddagger$} & \best{100.0} & \best{100.0} & \second{96.3} & \best{100.0} & \best{100.0} & \best{100.0} & \second{99.3} & \best{100.0} & \best{100.0} & \best{100.0} & \best{100.0} & \best{100.0} & \best{100.0} & \best{100.0} & \best{\textbf{99.7}} \\
\bottomrule
\end{tabular}
\label{tab:alfworld}
\end{table}


\section{Experiments}
\label{sec:experiments}
In this section, we evaluate vSFT through the following research questions motivated by our work.
\begin{enumerate}[leftmargin=*, topsep=0.1em, noitemsep]
\item Can vSFT outperform in-context learning counterparts and reinforcement learning methods while yielding more effective self-improvement than in-context reflection?
\item Can vSFT enable a weaker agent to match or surpass the performance of a stronger supervisor while remaining robust to supervisor choice across model families?
\item Can meta-training make the hypernetwork more resistant to overfitting under extended training and enable persistently internalized verbal feedback that generalizes to new scenarios?
\end{enumerate}
We next introduce the experimental setup and implementation details, and then present and analyze our findings, highlighting how they collectively address these questions.

\subsection{Experimental setup}

\myparagraph{Benchmarks}
Our experiments are conducted on two agent benchmarks: ALFWorld~\citep{ALFWorld20} and WebShop~\citep{yao2022webshop}.
ALFWorld is a text-based embodied environment where an agent completes long-horizon household instructions.
We report results on both its in-distribution (seen) and out-of-distribution (unseen) splits.
WebShop is a simulated e-commerce environment where agents navigate a web interface to find and purchase products satisfying user requirements.
We also evaluate on JSONSchemaBench~\citep{geng2025jsonschemabench},
a benchmark for assessing the ability of LLMs to generate JSON outputs that satisfy specified schemas.
It provides a clean testbed for comparing the effectiveness of in-context learning and our in-weight method in steering model behavior.
We use its GitHub Easy, Medium, and Hard subsets, which contain real-world JSON schemas of increasing size and structural complexity.

\myparagraph{Baselines}
We compare our vSFT against the in-context learning counterparts, including ReAct~\citep{yao2023react} and Reflexion~\citep{shinn2023reflexion}, conventional SFT, and policy-gradient reinforcement learning methods, including GRPO~\citep{shao2024deepseekmath}, GiGPO~\citep{feng2026group}, and SkillRL~\citep{xia2026skillrl}.
All results of the baselines are obtained under a consistent experimental setting.\footnote{We implement RL methods using Agent-R1~\citep{cheng2025agent}, so the results may differ substantially from those from \cite{feng2026group}.}
%

\myparagraph{Implementation details}
We use \texttt{Qwen3-\allowbreak{}4B-\allowbreak{}Instruct-\allowbreak{}2507} (abbreviated as Qwen3-4B) as the agent and \texttt{Qwen-\allowbreak{}Plus-\allowbreak{}2507} (abbreviated as Qwen-Plus) either as an alternative supervisor or as both the agent and supervisor to generate hypernetwork training data.
The hypernetwork follows the architecture of Doc-to-LoRA~\citep{charakorn2026doc} and is initialized with its pretrained weights.\footnote{ Directly using the pretrained hypernetwork yields near-zero success rates, highlighting the need for our proposed meta-training in \S\ref{sec:hypernetwork}. }
It has only 421M trainable parameters, and generates rank-8 LoRA updates applied to the MLP down-projection weights.
We finetune the hypernetwork using a learning rate of $4\times10^{-6}$.
For methods that use trial and error, we allow up to 10 trials per task.
Additional implementation details are provided in Appendix~\ref{app:details},
while further experimental results are provided in Appendix~\ref{app:exp}.

\begin{table}[hb!]
\centering
\scriptsize
\begin{minipage}[b]{0.51\linewidth}
\centering
\captionof{table}{
\small
\myparagraph{Comparisons on WebShop}
We compare vSFT against ICL and RL methods across different base models, reporting average task score and success rate.
}
\vspace{-1em}
\label{tab:webshop}
\begin{tabular*}{\linewidth}{l@{\extracolsep{\fill}}c@{\extracolsep{0pt}\hspace{1.7em}}l}
\toprule
Method & Score & SR \\
\midrule
\multicolumn{3}{l}{\textcolor{black!60}{\itshape Qwen3.8-Max}} \\
\addlinespace[0.15em]
\hspace*{0.5em}ReAct & 53.67 & 36.60 \\
\hspace*{0.5em}Reflexion & \third{79.16} & \second{59.80} \\
\addlinespace[0.3em]
\multicolumn{3}{l}{\textcolor{black!60}{\itshape Qwen-Plus-2507}} \\
\addlinespace[0.15em]
\hspace*{0.5em}ReAct & 51.66 & 32.80 \\
\hspace*{0.5em}Reflexion & 68.62 & 47.20 \\
\addlinespace[0.3em]
\multicolumn{3}{l}{\textcolor{black!60}{\itshape Qwen3-4B-Instruct-2507}} \\
\addlinespace[0.15em]
\hspace*{0.5em}ReAct & 40.74 & 18.80 \\
\hspace*{0.5em}Reflexion & 59.65 & \underline{30.80} \\
\hspace*{0.5em}Reflexion\smash{$^\dagger$} & 70.32 & \uuline{44.80} \\
\hspace*{0.5em}GRPO & 65.83 & 44.20 \\
\hspace*{0.5em}GiGPO & 67.13 & 50.00 \\
\addlinespace[0.1em]
\cdashline{1-3}
\addlinespace[0.2em]
\hspace*{0.5em}\textbf{vSFT} (one trial) & 49.64 & 24.45 \\
\hspace*{0.5em}\textbf{vSFT} & 73.55 & \underline{49.00} {\tiny(+18.2)} \\
\hspace*{0.5em}\textbf{vSFT}\smash{$^\dagger$} & \second{82.19} & \third{\uuline{59.80}} {\tiny(+15.0)} \\
\hspace*{0.5em}\textbf{vSFT}\smash{$^\ddagger$} & \best{83.58} & \best{62.00} \\
\bottomrule
\end{tabular*}
\end{minipage}
\hfill
\begin{minipage}[b]{0.45\linewidth}
\centering
\captionof{table}{
\small
\myparagraph{Ablation study on supervisor choice}
We evaluate vSFT with different supervisors on WebShop to assess its robustness to its choice.
}
\vspace{-1em}
\label{tab:ablation}
\begingroup
\setlength{\tabcolsep}{0.85em}
\begin{tabularx}{\linewidth}{ >{\raggedright\arraybackslash}X c c }
\toprule
Supervisor & Score & SR \\
\midrule
\multicolumn{3}{l}{\textcolor{black!60}{\itshape Qwen}} \\
\addlinespace[0.15em]
\hspace*{0.5em}Qwen3.8-Max & 76.47 & 53.00 \\
\hspace*{0.5em}Qwen3.7-Plus & 76.99 & 54.40 \\
\hspace*{0.5em}Qwen-Plus-2507 & 74.89 & 50.40 \\
\hspace*{0.5em}Qwen3-32B & 75.89 & 51.80 \\
\rowcolor{gray!13}
\hspace*{0.5em}Qwen3-4B-Instruct-2507 (baseline) & 73.55 & 49.00 \\
\addlinespace[0.30em]
\multicolumn{3}{l}{\textcolor{black!60}{\itshape GPT}} \\
\addlinespace[0.15em]
\hspace*{0.5em}GPT-5.6-Sol & 77.14 & 54.00 \\
\hspace*{0.5em}GPT-5.6-Terra & 78.35 & 55.40 \\
\hspace*{0.5em}GPT-5.6-Luna & 73.81 & 51.20 \\
\addlinespace[0.30em]
\multicolumn{3}{l}{\textcolor{black!60}{\itshape DeepSeek}} \\
\addlinespace[0.15em]
\hspace*{0.5em}DeepSeek-V4-Pro & 76.54 & 53.80 \\
\hspace*{0.5em}DeepSeek-V4-Flash & 76.49 & 53.60 \\
\addlinespace[0.30em]
\multicolumn{3}{l}{\textcolor{black!60}{\itshape GLM}} \\
\addlinespace[0.15em]
\hspace*{0.5em}GLM-5.2 & 76.38 & 52.40 \\
\hspace*{0.5em}GLM-5.1 & 78.22 & 56.00 \\
\bottomrule
\end{tabularx}
\endgroup
\end{minipage}
\end{table}

\subsection{vSFT enables effective self-improvement}

\myparagraph{Performance across methods}
As shown in Tabs.~\ref{tab:alfworld} and \ref{tab:webshop}, our vSFT outperforms Reflexion~\citep{shinn2023reflexion}, its vRL counterpart, by 13.9\% and 18.2\% (13.2\% and 15.0\% with a more capable supervisor) in average success rate across ALFWorld's two splits and success rate on Webshop, respectively.
Our vSFT also outperforms GRPO by 8.7\% and 4.8\% on ALFWorld and Webshop, while using less than 4\% and 10\% of GRPO's training rollout budget, respectively.
In Tab.~\ref{tab:withRL}, we observe that such advantage over RL persists under matched retry budgets,
and that the hypernetwork trained for the base model can be directly applied to its RL model, yielding greater performance gains than Reflexion.

\begin{figure}[t!]
\centering
\hspace{-4pt}
\includegraphics[width=0.98\columnwidth]{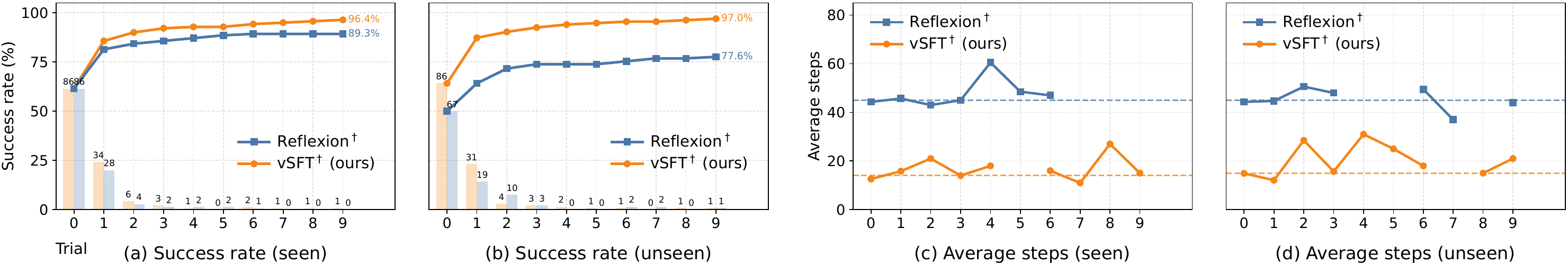}
\vspace{-1em}
\vspace{+4pt}
\caption{
\small
\myparagraph{Success rate and average steps across trials on ALFWorld}
The bars in panels (a) and (b) indicate the number of newly solved tasks per trial.
vSFT consistently improves over Reflexion while using shorter contexts and fewer steps, indicating more effective self-improvement and more efficient execution.
}
\label{fig:across_trial}
\end{figure}

\begin{figure}[t!]
\centering
\hspace{-4pt}
\includegraphics[width=0.98\columnwidth]{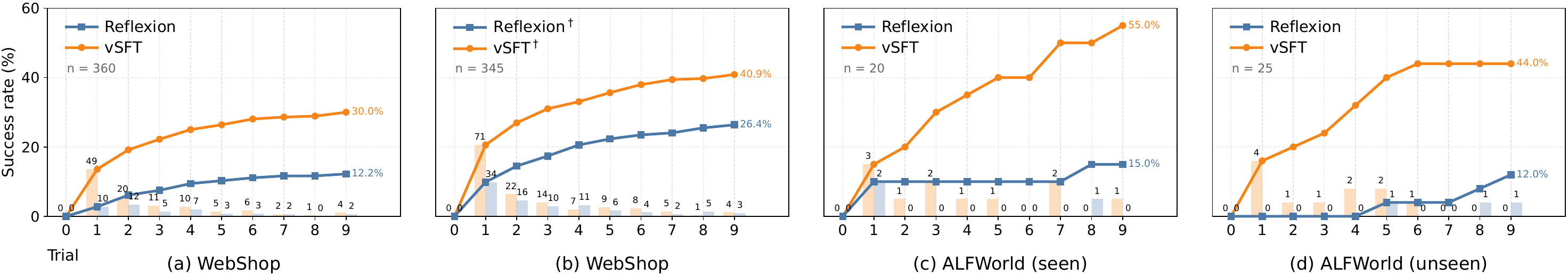}
\vspace{-1em}
\vspace{+4pt}
\caption{
\small
\myparagraph{Success rate on the shared initial failure set across trials}
We compare the self-improvement capabilities of vSFT and Reflexion by tracking their success rates across subsequent trials on the shared task set that both methods fail in the initial trial.
The figure indicates the set size as $n$.
}
\label{fig:common_failure}
\end{figure}

\myparagraph{Self-improvement across trials}
As shown in Tabs.~\ref{tab:alfworld} and \ref{tab:webshop}, our vSFT lags substantially behind most baselines in the initial trial, suggesting that its improved performance stems from more effective self-improvement across subsequent trials.
In Fig.~\ref{fig:across_trial}, we compare our vSFT with Reflexion in terms of success rate and average steps of completed tasks on ALFWorld.
We observe that vSFT consistently solves more previously failed tasks per trial while taking fewer average steps, suggesting that the agent improves itself more effectively and decisively via the hypernetwork in response to verbal feedback.
We further validate this conclusion in Fig.~\ref{fig:common_failure} by directly comparing the success rates across trials on the shared initial failure set.
It shows that vSFT still substantially outperforms Reflexion when we control for differences in the initial trial.

\begin{figure}[ht]
\centering
\hspace{-1cm}
\includegraphics[width=0.68\linewidth]{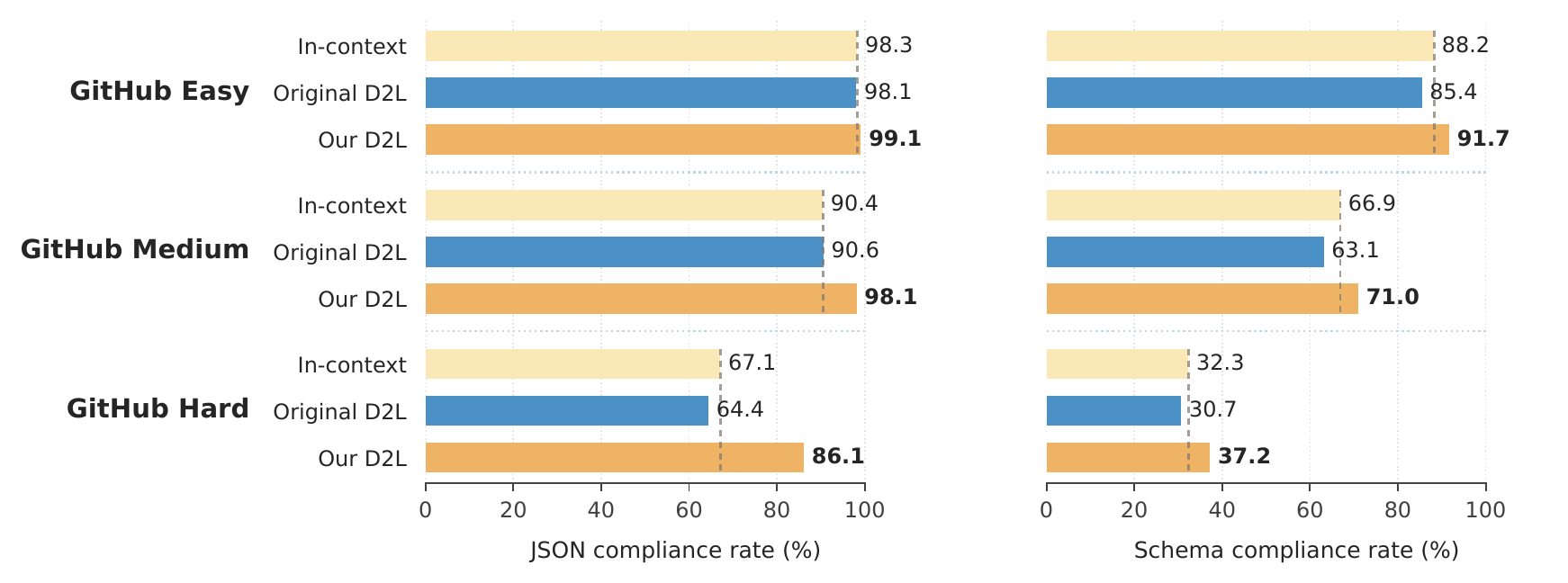}
\vspace{-1em}
\caption{%
\myparagraph{Comparisons on JSONSchemaBench}
We compare our hypernetwork, meta-trained on ALFWorld, against in-context learning and the original hypernetwork trained for document internalization across the GitHub Easy, Medium, and Hard splits of JSONSchemaBench, using Qwen3-4B-Instruct-2507.
The model's compliance with both JSON syntax and the specified JSON schema is reported.}
\label{fig:jsonschemabench}
\end{figure}

\myparagraph{Behavior steering}
In Fig.~\ref{fig:jsonschemabench}, we evaluate our meta-trained hypernetwork on JSONSchemaBench to isolate and assess its capability to steer model behavior.
%
We observe that our hypernetwork outperforms both the in-context baseline and the off-the-shelf hypernetwork from \citet{charakorn2026doc}, particularly on the hard split. 
These results further support our motivation that in-context learning is less effective for behavior steering, and suggest that merely internalizing textual content for question answering is also insufficient.

\subsection{vSFT learns beyond the supervisor}

\myparagraph{Performance across models}
As shown in Tabs.~\ref{tab:alfworld} and \ref{tab:webshop}, when the supervisor is a more capable model, Qwen-Plus, than the agent, Qwen3-4B, our vSFT outperforms the Reflexion baseline that uses Qwen-Plus-2507 as both the agent and supervisor, by 1.5\% and 12.6\% on ALFWorld and WebShop, respectively.
Moreover, when the hypernetwork is further trained solely on the data generated by Qwen-Plus, vSFT substantially outperforms this baseline, performs on par with Qwen3.8-Max on ALFWorld, and even surpasses it by 2.2\% on WebShop.

\myparagraph{Supervisor compatibility}
As discussed in \S\ref{sec:vsft}, compared to OPD \citep{li2026rethinking}, our vSFT is compatible with closed-source API models, as it merely requires feedback in natural language, which may also make it robust to supervisor choice.
In Tab.~\ref{tab:ablation}, we evaluate vSFT with supervisors from four distinct model families and observe that its performance can be further improved with more capable supervisors, even when the agent and supervisor belong to different model families.
Nonetheless, we also find that a more capable supervisor does not always yield better performance.

\vspace{+1em}

\begin{figure}[ht]
\centering
\includegraphics[width=0.9\textwidth]{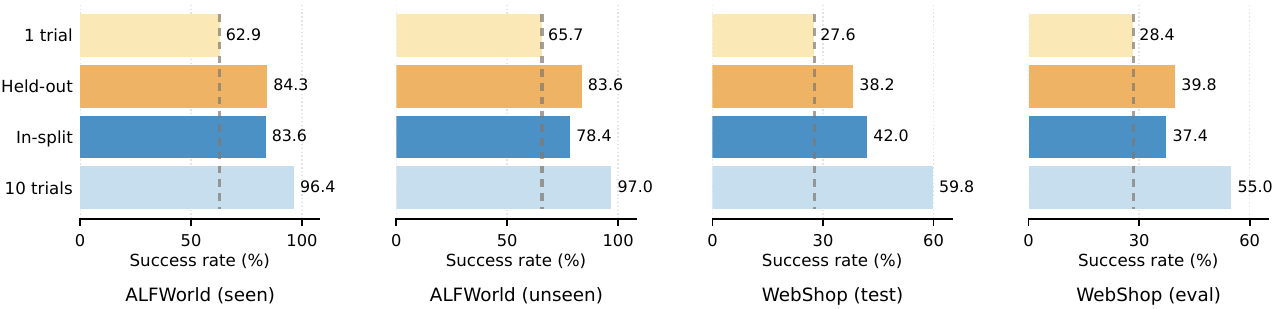}
\vspace{-1em}
\vspace{+4pt}
\caption{
\small
\myparagraph{Task-agnostic and persistent improvement}
Unlike our default setting, where each failed task has a separate LoRA adapter, we adapt the agent using a single LoRA adapter obtained by merging task-specific adapters collected from one of four splits: ALFWorld's seen and unseen splits and WebShop's evaluation and test splits.
The adapted agent is evaluated either on the same split used for merging (in-split) or on the other split of the same benchmark (held-out).
We include standard vSFT with 1 trial as a baseline and 10 trials as an upper bound.
All results are obtained under the vSFT$^\dagger$ setting.
}
\label{fig:lora_merge_generalization}
\end{figure}

\subsection{vSFT yields persistent and generalizable internalization}

\begin{wrapfigure}[10]{r}{0.4\textwidth}
\vspace{-\intextsep}
\centering
\includegraphics[width=0.35\textwidth]{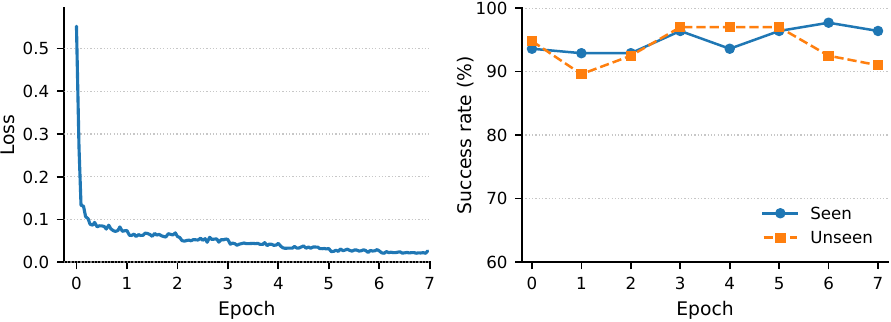}
\vspace{-0.5em}
\caption{
\small
\myparagraph{Training loss and success rates across epochs}
We train the hypernetwork on ALFWorld's training split (left panel) and evaluate on the two validation splits (right panel).
}
\label{fig:robust}
\end{wrapfigure}

\leavevmode
\myparagraph{Training robustness}
As discussed in \S\ref{sec:hypernetwork}, the hypernetwork is trained to learn how to adapt the agent rather than directly produce an adapted model.
We examine whether this meta-training process preserves generalization under extended and excessive training.
As shown in Fig.~\ref{fig:robust}, although the training loss consistently decreases over additional epochs, vSFT's success rates do not collapse on either validation split, even on the out-of-distribution split.
The hypernetwork thus shows no signs of severe overfitting within the evaluated training epochs.

\myparagraph{Cross-task improvement}
Our vSFT currently provides specific verbal feedback for each failed task and thus generates separate LoRA adapters.
Although these adapters can be viewed as fast weights~\citep{wen2026mixture,zhang2026fast} or managed using certain infrastructure~\citep{lab2026mint}, we investigate the potential of using a single model across all tasks that achieves persistent and generalizable self-improvement.
In Fig.~\ref{fig:lora_merge_generalization}, we adapt the agent using a LoRA adapter that merges adapters collected from one split and evaluate its performance on another split.
The results show that this single adapted model still substantially outperforms the one-trial vSFT baseline, suggesting that internalized verbal feedback can generalize to new tasks in a parametric form.
By contrast, additional results in Appendix~\ref{app:exp} indicate that such improvements are difficult to achieve through in-context verbal feedback.
\section{Conclusion and future work}
We have proposed a novel test-time adaptation method for LLM agents, vSFT, which provides a learnable and effective self-improvement mechanism by iteratively internalizing verbal feedback in a closed loop.
We have conducted extensive experiments on ALFWorld and WebShop.
The empirical results confirmed that a trained hypernetwork could effectively internalize verbal feedback into an agent, yielding significantly improved performance.
Nevertheless, in the current vSFT framework, the hypernetwork that generates LoRA adapters still remains a separate component from the agent.
Thus, exploring native self-improvement models that directly generate fast weights to adapt themselves for potentially improving both efficiency and accuracy is worth future work.

\clearpage
\begingroup
\setlength{\bibsep}{4pt}
\bibliographystyle{plainnat}
\bibliography{references}
\endgroup
\beginappendix
\section{A parameter-space perspective}
\begin{figure}[h!]
\centering
\hspace{+3pt}
\includegraphics[width=0.6\textwidth]{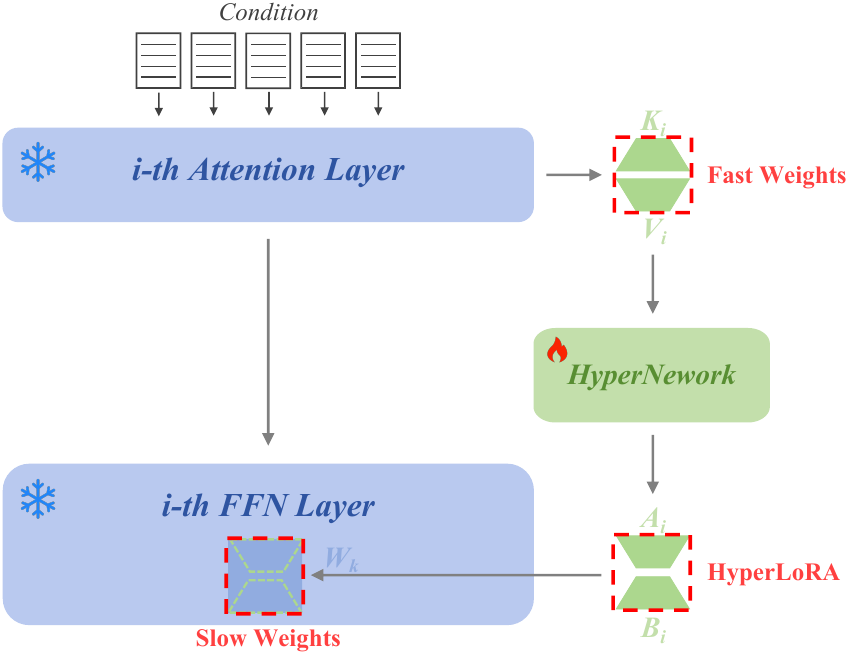}
\vspace{-1em}
\vspace{+2pt}
\caption{
\myparagraph{Illustration of hypernetwork-based parameter adaptation}
The textual content to be internalized is first processed by the base model in a single forward pass to collect its key-value states in attention layers.
The hypernetwork then uses these states to generate LoRA adapters, which are applied to the down-projection weight matrices of the FFN layers.
}
\label{fig:hypernetwork}
\end{figure}

To offer some intuition for
how natural language can be mapped into parameter space,
we draw on the concept of fast weights.
%
Fig.~\ref{fig:hypernetwork} illustrates the hypernetwork workflow we adopt, following the design proposed in Doc-to-LoRA~\citep{charakorn2026doc}.

%
\citet{wen2026mixture} reinterpret attention as a two-layer fast-weight MLP,
whose hidden dimension equals the sequence length
and whose weights are formed by key-value states.
In conventional architectures,
fast weights in attention layers
and pretrained slow weights (i.e., model parameters) in FFN layers
play distinct roles and remain largely separate.
As illustrated in Fig.~\ref{fig:hypernetwork},
the hypernetwork bridges these two forms of weights
by mapping KV states into low-rank updates
to slow weights.
This provides a fast-to-slow pathway for internalizing working memory into model parameters.

%

\section{Additional implementation details}
\label{app:details}

\myparagraph{vSFT variants}
As shown in Tab.~\ref{tab:alfworld}, we implement three vSFT variants that differ in the agent and supervisor models used during data curation and evaluation.
For clarity, we summarize their differences in Tab.~\ref{tab:vsft_variants}.

\begin{table}[h]
\centering
\small
\caption{
\small
Agent and supervisor models across vSFT variants during data curation and evaluation.
}
\vspace{-1em}
\vspace{+1pt}
\begin{tabular}{lcccc}
\toprule
\multirow{2}{*}{Method} & \multicolumn{2}{c}{Data Curation} & \multicolumn{2}{c}{Evaluation} \\
\cmidrule(lr){2-3}
\cmidrule(lr){4-5}
& Agent & Supervisor & Agent & Supervisor \\
\midrule
vSFT & \cellcolor{gray!15}Qwen3-4B & \cellcolor{gray!15}Qwen3-4B & \cellcolor{gray!15}Qwen3-4B & \cellcolor{gray!15}Qwen3-4B \\
vSFT$^{\dagger}$ & \cellcolor{gray!15}Qwen3-4B & Qwen-Plus & \cellcolor{gray!15}Qwen3-4B & Qwen-Plus \\
vSFT$^{\ddagger}$ & Qwen-Plus & Qwen-Plus & \cellcolor{gray!15}Qwen3-4B & Qwen-Plus \\
\bottomrule
\end{tabular}
\vspace{-3pt}
\label{tab:vsft_variants}
\end{table}

\myparagraph{LoRA merging}
For the experiments in Fig.~\ref{fig:lora_merge_generalization}, we merge the effective weight updates (defined in Eq.~\ref{eq:lora_effective_weight}) of all LoRA adapters collected from successful tasks using a weighted variant of TIES-Merging~\citep{yadav2023ties}.
We assign equal total weight to each of the six task categories in ALFWorld and equal weight to each unique verbal feedback document in WebShop.

\myparagraph{Rollout cost}
For data curation, we collect the trajectories described in Eq.~\ref{eq:meta_dataset} from the training splits of ALFWorld and WebShop under the Reflexion (i.e., vRL) framework.
In total, this results in 10,712 rollouts on ALFWorld and 86,533 rollouts on WebShop.
For comparison, the default GRPO training recipe in Agent-R1~\cite{cheng2025agent} requires 284,160 rollouts on ALFWorld and 926,720 rollouts on WebShop.\footnote{ These results are computed as $\lfloor N/16 \rfloor \times 16 \times 8 \times 10$, where $N$ denotes the training-set size (3,553 for ALFWorld and 11,587 for WebShop), 16 is the batch size, 8 is the group size, and 10 is the number of epochs.}
Our data curation therefore requires only 3.8\% and 9.3\% of GRPO's rollout budgets on ALFWorld and WebShop, respectively.

\myparagraph{Inference cost}
Our work targets more effective behavior steering rather than inference efficiency.
Compared with Reflexion, vSFT reduces context usage and steps (Fig.~\ref{fig:across_trial}) but adds overhead for generating and loading LoRAs.
On ALFWorld's seen split, using a single RTX 4090 (48~GB) GPU, Reflexion takes 592~s, while vSFT takes 608~s (3\% longer). LoRA generation and loading account for 61~s (10\%) and 4~s (1\%), respectively.

\myparagraph{Training cost}
Compared with GRPO, vSFT has a simpler training pipeline and requires substantially less training time.
vSFT decouples data collection from optimization, allowing offline and off-policy data collection.
Moreover,
the hypernetwork is trained with standard cross-entropy loss on previously collected successful trajectories, avoiding online reward evaluation, advantage estimation, and policy-ratio clipping during optimization.
On ALFWorld, using a single RTX 6000D (84~GB) GPU, GRPO takes 62 hours per epoch for 10 epochs, whereas vSFT takes 9 hours per epoch for 5 epochs, plus 8 hours for data collection.
Overall, vSFT reduces the total training time from 623 to 54 hours, achieving a 91\% reduction.

\begin{figure}[ht]
\centering
\includegraphics[width=0.65\textwidth]{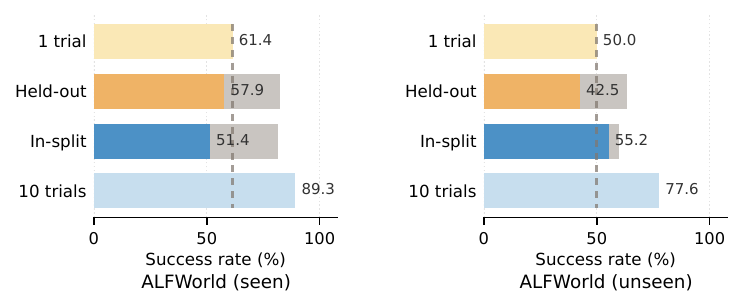}
\vspace{-1em}
\caption{
\small
\myparagraph{Persistent skill bank across tasks}
We use Qwen-Plus to summarize task-specific verbal feedback into a persistent skill bank for all tasks.
We evaluate the agent with this bank on either the same split used to construct it (in-split) or the other split (held-out).
We include Reflexion with 1 trial as a baseline and with 10 trials as an upper bound.
These results are obtained under the Reflexion$^\dagger$ setting.
}
\label{fig:context_transfer}
\end{figure}

\section{Additional experimental results}
\label{app:exp}

\myparagraph{In-context transfer}
In Fig.~\ref{fig:lora_merge_generalization}, we show that internalized task-specific verbal feedback can be consolidated through LoRA merging into a persistent, task-agnostic parametric improvement to the agent (in-parameter transfer).
We now investigate whether in-context verbal feedback can achieve a similar effect by having an LLM summarize it into a skill bank that the agent persistently retains in context across tasks.
As shown in Fig.~\ref{fig:context_transfer}, this in-context counterpart struggles to outperform the baseline and yields substantially smaller gains over the baseline than its in-parameter counterpart, with the gaps highlighted by the grey bars.

\myparagraph{Additional performance comparisons}
In Tab.~\ref{tab:additional_alfworld}, we further compare with several methods whose ALFWorld results are only partially reported and thus cannot be included in the comprehensive comparison in Tab.~\ref{tab:alfworld}.

\begin{table}[ht]
\centering
\scriptsize
\setlength{\tabcolsep}{8pt}
\caption{
\small
\myparagraph{Additional comparisons on ALFWorld}
MemRL and AdaMEM are two methods that enhance the agent through external memory, while ReflAct does so through real-time reflections.
}
\label{tab:additional_alfworld}
\vspace{-1em}
\begin{tabular}{lcc}
\toprule
Method & Seen & Unseen \\
\midrule
\multicolumn{3}{l}{\textcolor{black!60}{\itshape GPT-5-mini}} \\
\addlinespace[0.15em]
\hspace*{0.5em}MemRL~\citep{zhang2026memrl} & 97.9 & -- \\
\addlinespace[0.3em]
\multicolumn{3}{l}{\textcolor{black!60}{\itshape GPT-4o}} \\
\addlinespace[0.15em]
\hspace*{0.5em}ReflAct~\citep{kim2025reflact} & -- & 93.3 \\
\addlinespace[0.3em]
\multicolumn{3}{l}{\textcolor{black!60}{\itshape Qwen3-8B}} \\
\addlinespace[0.15em]
\hspace*{0.5em}LatentSkill~\citep{yu2026latentskill} & 74.3 & 69.4 \\
\addlinespace[0.3em]
\multicolumn{3}{l}{\textcolor{black!60}{\itshape Qwen3-4B-Instruct-2507}} \\
\addlinespace[0.15em]
\hspace*{0.5em}AdaMEM-MAX~\citep{zhang2026adamem} & 65.5 & -- \\
\addlinespace[0.1em]
\cdashline{1-3}
\addlinespace[0.2em]
\hspace*{0.5em}\textbf{vSFT} & 92.9 & 92.5 \\
\hspace*{0.5em}\textbf{vSFT}\smash{$^\ddagger$} & 99.3 & 100.0 \\
\bottomrule
\end{tabular}
\end{table}

\begin{figure}[h]
    \centering
    \captionsetup[subfigure]{
        font=small,
        labelfont=normalfont,
        labelformat=parens,
        skip=-8pt
    }
    \captionsetup{subrefformat=parens}

    \begin{subfigure}[t]{0.42\linewidth}
        \centering
        \includegraphics[width=\linewidth]{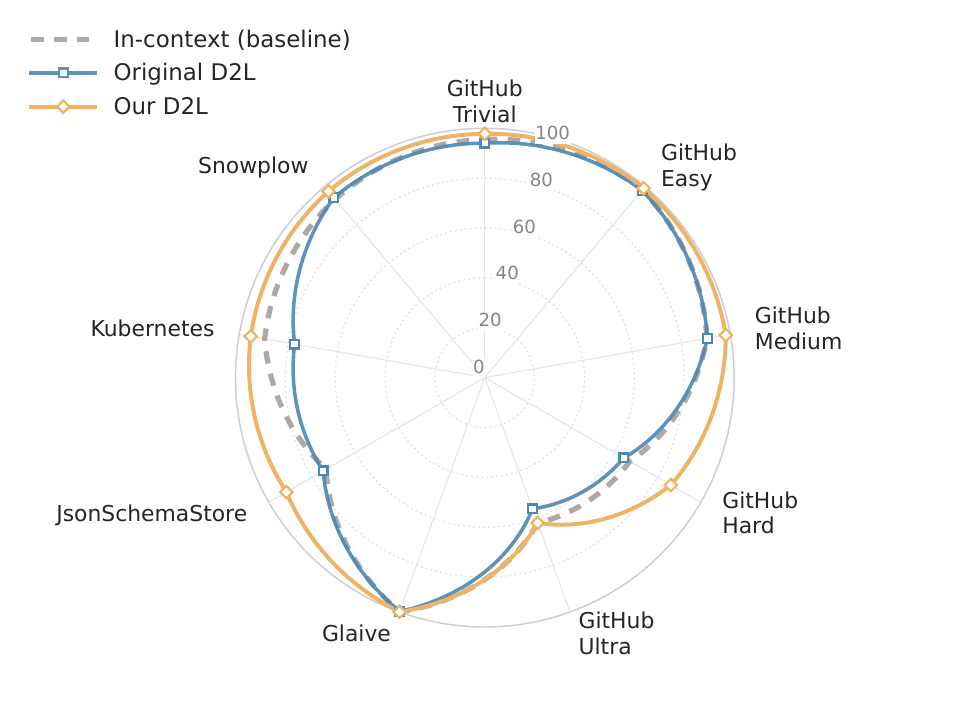}
        \caption{JSON compliance rate (\%).}
        \label{fig:jsonschemabench-json-raw}
    \end{subfigure}
    \begin{subfigure}[t]{0.42\linewidth}
        \centering
        \includegraphics[width=\linewidth]{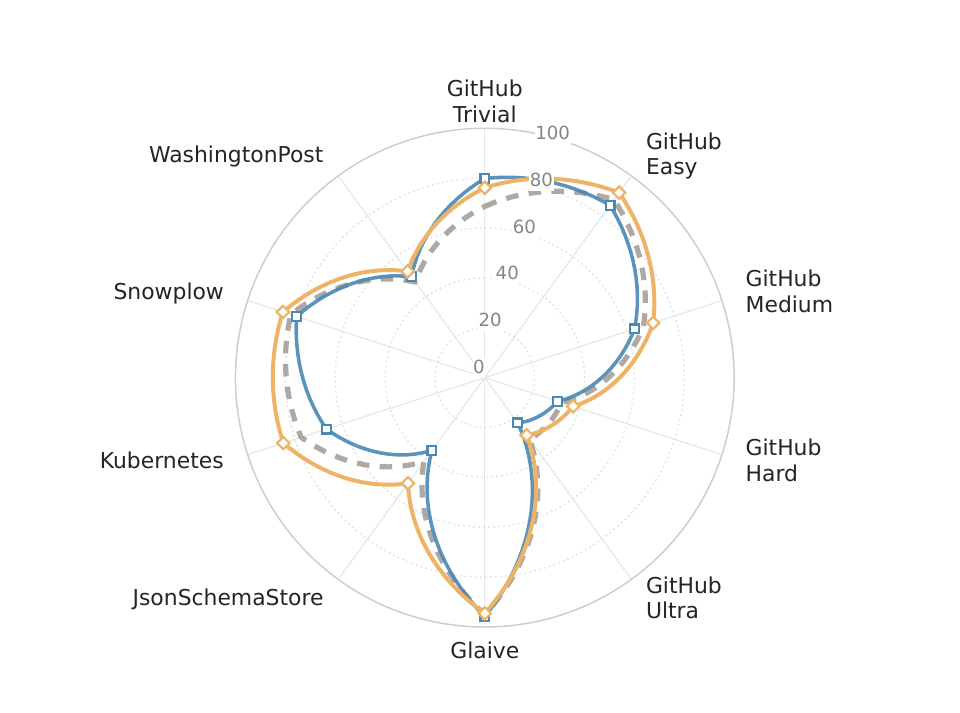}
        \caption{Schema compliance rate (\%).}
        \label{fig:jsonschemabench-schema-raw}
    \end{subfigure}


    \begin{subfigure}[t]{0.42\linewidth}
        \centering
        \includegraphics[width=\linewidth]{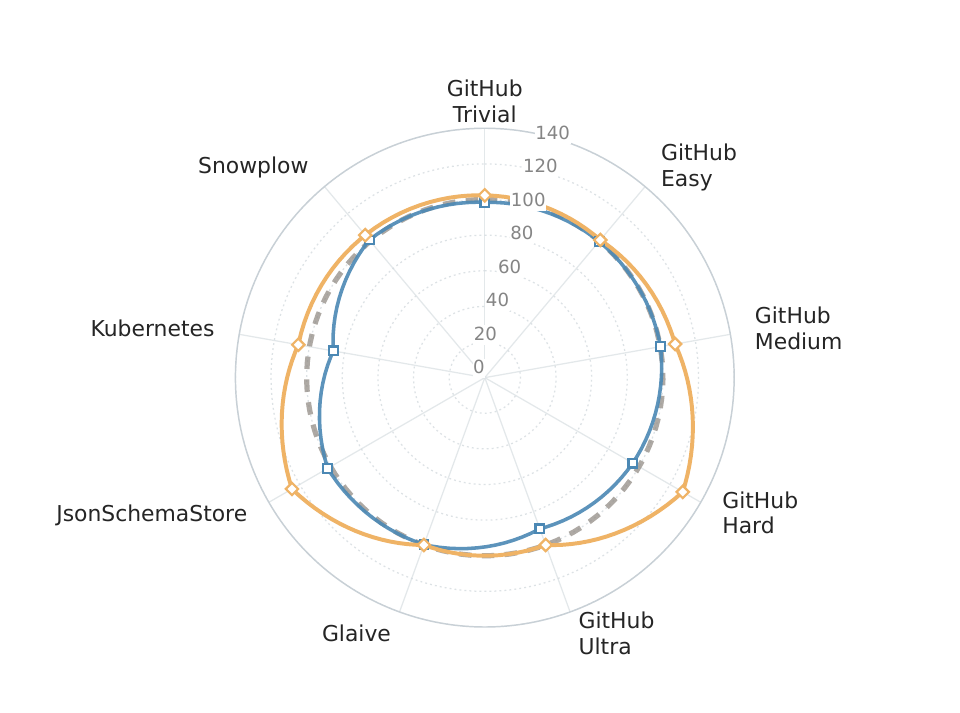}
        \caption{JSON compliance / baseline (\%).}
        \label{fig:jsonschemabench-json-normalized}
    \end{subfigure}
    \begin{subfigure}[t]{0.42\linewidth}
        \centering
        \includegraphics[width=\linewidth]{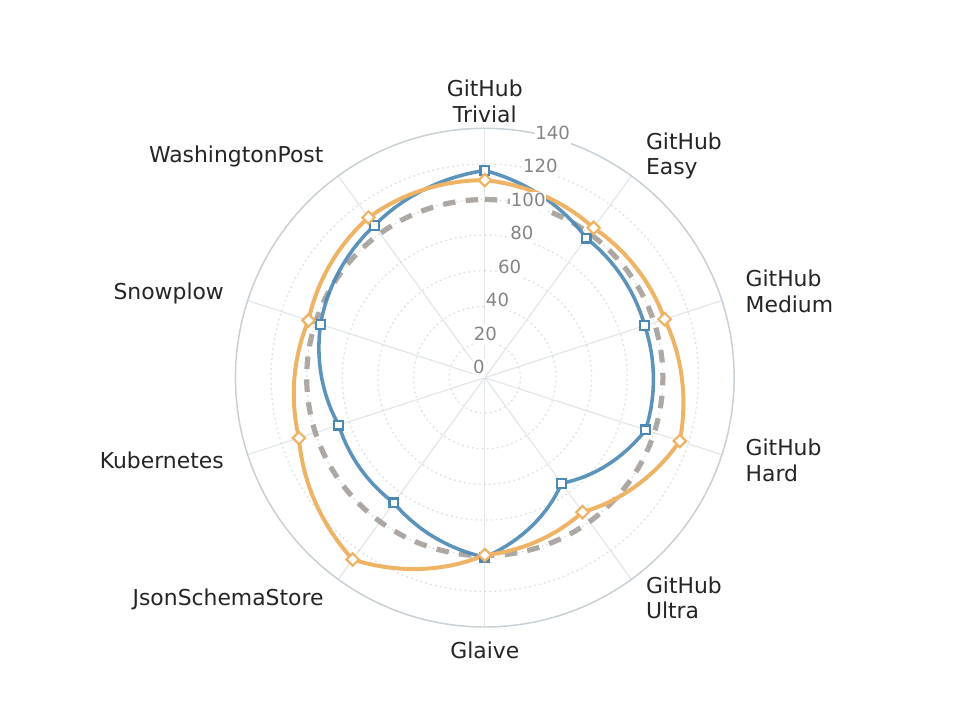}
        \caption{Schema compliance / baseline (\%).}
        \label{fig:jsonschemabench-schema-normalized}
    \end{subfigure}
\vspace{-2pt}
\caption{
\myparagraph{Additional comparisons on JSONSchemaBench}
Consistent with Fig.~\ref{fig:jsonschemabench}, we report results on all splits of JSONSchemaBench. The raw compliance rates with JSON syntax and the specified JSON schema are reported in panels~\subref{fig:jsonschemabench-json-raw} and~\subref{fig:jsonschemabench-schema-raw}, respectively, while the relative performance against the in-context baseline is reported in panels~\subref{fig:jsonschemabench-json-normalized} and~\subref{fig:jsonschemabench-schema-normalized}.
}
\label{fig:jsonschemabench-full}
\end{figure}

\myparagraph{More JSONSchemaBench results}
In Fig.~\ref{fig:jsonschemabench-full},
we observe that our hypernetwork-generated LoRAs consistently outperform the in-context baseline across all splits of JSONSchemaBench.
This advantage is particularly pronounced on more challenging splits, such as GitHub Hard and JSONSchemaStore.

\myparagraph{KV cache resue}
Our vSFT generates and equips LoRA adapters at runtime, such that the model parameters may change during inference.
While such dynamic parameter adaptation provides greater model flexibility and plasticity and enables more reliable behavior steering, it also raises practical concerns for inference systems, particularly regarding KV cache reuse.
Specifically, KV states may be computed and cached under one LoRA adapter, while subsequent inference proceeds after that adapter has been replaced or detached, resulting in a mismatch between the parameters used to construct the cache and those used for subsequent decoding.
To investigate the impact of this mismatch, we conduct experiments on SCBench~\citep{li2025scbench}, a benchmark designed to evaluate long-context generation under KV cache reuse.
As shown in Tab.~\ref{tab:scbench}, 
although the KV cache is computed with a LoRA adapter generated specifically for ALFWorld, 
the resulting KV states can still transfer to other tasks and even yield performance gains.

\definecolor{SCBDelta}{HTML}{F4F6F8}
\definecolor{SCBMean}{HTML}{EAF0F5}
\begin{table}[bht]
\centering
\caption{
\myparagraph{Results on SCBench}
We report SCBench results under different inference models and KV cache sources, 
and the performance changes induced by switching the cache source.
Base inference
denotes inference with the base model, while
LoRA inference
denotes inference with the model equipped with a LoRA adapter; 
base KV
and 
LoRA KV
are defined similarly.
%
The LoRA adapter used in these experiments is generated by our hypernetwork meta-trained on ALFWorld, 
internalizing on the system prompt and few-shot examples.
}
\label{tab:scbench}
\small
\vspace{-1em}
\begin{tabular}{lrr>{\columncolor{SCBDelta}}r@{\hspace{14pt}}rr>{\columncolor{SCBDelta}}r}
\toprule
\multirow{2}{*}{Task} & \multicolumn{3}{c}{Base inference} & \multicolumn{3}{c}{LoRA inference} \\
\cmidrule(lr){2-4}\cmidrule(lr){5-7}
 & Base KV & LoRA KV & $\Delta$ & Base KV & LoRA KV & $\Delta$ \\
\midrule
Retr.KV & 54.6 & 62.2 & $+7.6$ & 13.2 & 12.4 & $-0.8$ \\
Retr.MultiHop & 40.0 & 32.0 & $-8.0$ & 40.3 & 19.5 & $-20.8$ \\
Code.RepoQA & 53.6 & 53.0 & $-0.7$ & 45.5 & 38.9 & $-6.6$ \\
En.QA & 27.3 & 24.5 & $-2.8$ & 27.3 & 24.8 & $-2.4$ \\
Zh.QA & 3.4 & 4.0 & $+0.6$ & 6.3 & 6.8 & $+0.5$ \\
En.MultiChoice & 17.8 & 29.7 & $+11.9$ & 67.6 & 64.3 & $-3.3$ \\
Math.Find & 28.3 & 14.2 & $-14.2$ & 17.3 & 6.5 & $-10.8$ \\
ICL.ManyShot & 70.7 & 70.7 & $0.0$ & 40.4 & 37.4 & $-3.0$ \\
En.Sum & 36.1 & 36.5 & $+0.4$ & 35.8 & 36.3 & $+0.5$ \\
Mix.Sum+NIAH (Sum) & 36.4 & 37.3 & $+0.9$ & 36.4 & 37.6 & $+1.2$ \\
Mix.Sum+NIAH (NIAH) & 65.0 & 83.6 & $+18.6$ & 98.2 & 90.7 & $-7.5$ \\
Mix.RepoQA+KV (RepoQA) & 60.2 & 54.6 & $-5.7$ & 49.7 & 44.3 & $-5.4$ \\
Mix.RepoQA+KV (KV) & 81.8 & 91.2 & $+9.4$ & 81.0 & 82.7 & $+1.7$ \\
\midrule
\rowcolor{SCBMean}
\textbf{Average} & 44.3 & 45.6 & $\mathbf{+1.4}$ & 43.0 & 38.6 & $\mathbf{-4.4}$ \\
\bottomrule
\end{tabular}
\end{table}

\myparagraph{Compatibility with RL}
Intuitively, vSFT can be combined with RL methods.
Considering that RL induces smaller distributional shifts from the base model than SFT,
we apply the hypernetwork trained for the base model to its RL model without further training. 
As shown in Tab.~\ref{tab:withRL},
vSFT still enables more effective improvement on top of RL.

\begin{table}[t]
\centering
\small
\caption{
\small
\myparagraph{Combining vSFT with RL}
We directly apply vSFT to a GiGPO-trained agent without further training the hypernetwork.
The results are reported on ALFWorld.
$\Delta N_{\mathrm{succ}}$ denotes the number of newly solved tasks relative to the baseline.
To mitigate randomness and narrow the inference budget gap between the baseline and reflection methods, 
we introduce an extended setting that allows three attempts at each of three temperatures (0, 0.1, and 0.2) and reduces the number of trials for reflection methods from 10 to 2.
}
\label{tab:withRL}
\vspace{-1em}
\vspace{+1pt}
\begin{tabular}{lcccc}
\toprule
\multirow{2}{*}{Method}
& \multicolumn{3}{c}{Success Rate (\%)}
& \multirow{2}{*}{$\Delta N_{\mathrm{succ}}$} \\
\cmidrule(lr){2-4}
& Seen & Unseen & Average & \\
\midrule
\multicolumn{5}{l}{\textit{1 temperature, 1 retry, 10 trials (default)}} \\
\addlinespace[2pt]
\rowcolor{gray!15}
GiGPO (baseline) & 80.0 & 71.6 & 75.8 & 0 \\
\hspace{1em}$+$ Reflexion
    & 80.7 & 75.4 & 78.1 & +6 \\
\hspace{1em}$+$ vSFT
    & 85.0 & 83.6 & 84.3 & +23 \\
\hspace{1em}$+$ vSFT$^{\dagger}$
    & 87.9 & 85.8 & 86.9 & +30 \\
\hspace{1em}$+$ vSFT$^{\ddagger}$
    & 89.3 & 89.6 & 89.5 & +37 \\
\midrule
\multicolumn{5}{l}{\textit{3 temperatures, 9 retries, 2 trials (extended)}} \\
\addlinespace[2pt]
GRPO & 85.0 & 86.6 & 85.8 & N/A \\
\rowcolor{gray!15}
GiGPO (baseline) & 87.9 & 84.3 & 86.1 & 0 \\
\hspace{1em}$+$ Reflexion
    & 87.9 & 85.1 & 86.5 & +1 \\
\hspace{1em}$+$ vSFT
    & 90.0 & 87.3 & 88.7 & +7 \\
\hspace{1em}$+$ vSFT$^{\dagger}$
    & 90.7 & 90.3 & 90.5 & +12 \\
\hspace{1em}$+$ vSFT$^{\ddagger}$
    & 90.7 & 91.0 & 90.9 & +13 \\
\bottomrule
\end{tabular}
\end{table}

\section{Agent system prompts}
We provide the system prompts used by the agent on ALFWorld and WebShop below.
As shown, our prompts are detailed and informative, providing sufficient guidance to the agent and thereby allowing the in-context baselines to realize their full potential.

\begin{vsftsystemprompt}[label={prompt:alfworld_actor_system}]{ALFWorld system prompt}
You are an ALFWorld agent. You interact with a household environment one step at a time.

STRICT OUTPUT RULES (you must follow every rule):
1. Output EXACTLY ONE line per turn. Never output multiple lines.
2. Your line must be ONE of the following:
   - A brief thought:  think: <your reasoning>
   - A valid ALFWorld action (see the complete list below)
3. NEVER output observations, environment feedback, or results. Only the environment provides observations — you never write them.

COMPLETE LIST OF VALID ACTIONS:
  go to {receptacle}              — move to a receptacle/location
  open {receptacle}               — open a closed container
  close {receptacle}              — close an open container
  take {object} from {receptacle} — pick up an object from a receptacle
  move {object} to {receptacle}   — place an object into/onto a receptacle
  put {object} in/into {target}   — place an object into another object or receptacle
  heat {object} with {receptacle} — heat an object using a microwave
  cool {object} with {receptacle} — cool an object using a fridge
  clean {object} with {receptacle} — clean an object using a sinkbasin
  slice {object} with {knife}     — slice an object using a knife
  use {object}                    — toggle an object on/off (e.g. desklamp)
  examine {receptacle or object}  — examine a receptacle or object in detail
  look                            — observe the current surroundings
  inventory                       — check what you are currently carrying

UNDERSTANDING ENVIRONMENT FEEDBACK:
- "OK." after a think: line means your thought was recorded. It is NOT asking you to repeat or elaborate — just proceed with your next action.
- "Nothing happens." means your action was invalid (wrong syntax, missing preconditions, or targeting a nonexistent object). When you see this, try a different action instead of repeating the same one.
\end{vsftsystemprompt}


\begin{vsftsystemprompt}[label={prompt:webshop_actor_system}]{WebShop system prompt}
You are a WebShop agent. Your goal is to find and purchase a product that perfectly matches a given instruction on a simulated e-commerce website.

VALID ACTIONS:
  search[keywords]  — search for products
  click[target]     — click any [bracketed item] visible on the current page. This includes product ASINs, option values, navigation buttons, and sub-page links.
  think[reasoning]  — record your internal reasoning (environment replies OK.)

SEARCH RULE (CRITICAL):
  search[keywords] is ONLY valid on the initial page (the page that shows "[Search]" and the instruction).
  If you are on ANY other page (search results, item page, or sub-page), you MUST first click[Back to Search] to return to the initial page before issuing search[].
  Issuing search[] from a non-initial page will result in "Invalid action!" and waste a step.

ENVIRONMENT CONSTRAINTS:
- Search results show at most 3 products per page. Use [Next >] and [< Prev] to browse additional pages if the first page has no suitable match.
- Products may have selectable options (e.g. color, size, flavor). Select all relevant options before clicking [Buy Now]; missing options will reduce your score.
- Use [< Prev] to navigate back from a sub-page to the item page, or from the item page to search results.

OUTPUT FORMAT:
1. Output EXACTLY ONE line per turn.
2. NEVER output observations or environment feedback — only the environment provides those.
\end{vsftsystemprompt}

\end{document}